\documentclass{article}

\usepackage{microtype}
\usepackage{graphicx}
\usepackage{subfigure}
\usepackage{booktabs} 

\usepackage{hyperref}

\usepackage{amsmath,amssymb}
\usepackage{mathtools}
\usepackage{color}
\usepackage{dsfont}
\usepackage{wrapfig}
\usepackage{graphicx}

\newcommand{\defeq}{\vcentcolon=}

\newcommand{\KL}{\text{KL}}
\newcommand{\Ent}{\text{H}}

\newcommand{\bbE}{\mathbb{E}}

\newcommand{\qq}{\pi}
\newcommand{\pp}{\pi_0}
\newcommand{\traj}{\tau}
\def\EE{\mathbb{E}}

\newcommand{\eg}{e.g.}

\newcommand{\hyeonwoo}[1]{{\color{blue}{HW: #1}}}

\newcommand{\nicolas}[1]{{\color{red}{nicolas: #1}}}

\newcommand{\appendixgeneralframework}{A??}

\usepackage[accepted]{icml2019}

\icmltitlerunning{Exploiting Hierarchy for Learning and Transfer in KL-regularized RL}

\begin{document}

\twocolumn[
\icmltitle{Exploiting Hierarchy for Learning and Transfer in KL-regularized RL}
\icmlsetsymbol{equal}{*}

\begin{icmlauthorlist}
\icmlauthor{Dhruva Tirumala}{equal,deepmind}
\icmlauthor{Hyeonwoo Noh}{equal,postech}
\icmlauthor{Alexandre Galashov}{deepmind}
\icmlauthor{Leonard Hasenclever}{deepmind}
\icmlauthor{Arun Ahuja}{deepmind}
\icmlauthor{Greg Wayne}{deepmind}
\icmlauthor{Razvan Pascanu}{deepmind}
\icmlauthor{Yee Whye Teh}{deepmind}
\icmlauthor{Nicolas Heess}{deepmind}
\end{icmlauthorlist}

\icmlaffiliation{postech}{Department of Computer Science and Engineering, POSTECH, Pohang, Korea}
\icmlaffiliation{deepmind}{DeepMind, London, UK}

\icmlcorrespondingauthor{Dhruva Tirumala}{dhruvat@google.com}
\icmlcorrespondingauthor{Hyeonwoo Noh}{shgusdngogo@postech.ac.kr}

\icmlkeywords{Machine Learning, ICML}

\vskip 0.3in
]



\printAffiliationsAndNotice{\icmlEqualContribution} 

\begin{abstract}
As reinforcement learning agents are  tasked with solving more challenging and diverse tasks, the ability to incorporate prior knowledge into the learning system and the ability to exploit reusable structure in solution space is likely to become increasingly important.
The KL-regularized expected reward objective constitutes a convenient tool to this end. It introduces an additional component, a default or prior behavior, which can be learned alongside the policy and as such partially transforms the reinforcement learning problem into one of behavior modelling.
In this work we consider the implications of this framework in case where both the policy and default behavior are augmented with latent variables. We discuss how the resulting hierarchical structures can be exploited to implement different inductive biases and how the resulting modular structures can be exploited for transfer. 
Empirically we find that they lead to faster learning and transfer on a range of continuous control tasks.
\end{abstract}

\section{Introduction}

\iftrue



Recent advances have greatly improved data efficiency, scalability, and stability of reinforcement learning  (RL) algorithms leading to successful applications in a number of domains ~\cite{heess2017emergence,riedmiller2018learning,andrychowicz2018learning,mnih2015human,OpenAI_dota,silver2016mastering}. 
%
Many problems, however, still remain challenging to solve or require large numbers of interactions with the environment; a situation that is likely to get worse as we attempt to tackle increasingly challenging and diverse problems.

A general avenue for reducing sample complexity in machine learning is the use of appropriate prior knowledge, or inductive biases. In RL, curricula, learning and transfer across task distributions, and the use of demonstrations have been shown to be powerful tools in this regard. 
The success of these ideas, and more generally our ability build agents that have a chance of succeeding in lifelong learning scenarios, relies on the ability to transfer and generalize behaviours across tasks, and thus on mechanisms for extracting knowledge from existing solutions and for using such knowledge to shape the solutions to new problems.

Recently \cite{teh2017distral,galashov2018information} have developed a framework for knowledge representation and transfer in RL that allows to express and exploit inductive biases in reinforcement learning problems.
The approach relies on a connection between the KL-regularized objective~\cite{todorov2007linearly,kappen2012optimal,rawlik2012stochastic,schulman2017equivalence} and probabilistic models. 
%
%
One way to think of a policy is that, together with the environment dynamics, it defines a distribution over trajectories, and the framework uses probabilistic models of trajectories to express prior knowledge about the solution distribution of a RL problem. 
To this end it introduces a second component, a prior or default behaviour, to which the policy is encouraged to remain close in terms of the Kullback-Leibler (KL) divergence. This prior can simplify the learning problem, for instance by reducing the trajectory space that needs to be searched, or by steering the learning algorithm away from undesirable solutions. Rather than manually crafting specific constraints which may or may not be appropriate for a given task or task distribution, it is possible to learn the prior from data, and \cite{teh2017distral,galashov2018information} have shown that this can be effective both in multi-task and transfer learning scenarios. \cite{galashov2018information} demonstrate how information asymmetry, i.e.\ restricting the prior's access to certain parts of the state space can be used to selectively transfer or generalize certain aspects of a learned behavior across tasks or different parts of the state space.

In this paper we extend and generalize this framework and study priors and policies that are hierarchically structured and augmented with latent variables. The introduction of latent variables can increase model capacity, and allow for richer classes of distributions (\eg~non-Gaussian). Importantly, it enables a broader range of inductive biases: It allows us to express more specific constraints on the information processing capabilities of the agent
leading to priors that have more diverse generalization properties. In particular, the hierarchical model structure allows us to mirror more closely the natural structure of many problems. For instance, in motor control, one tends to observe a separation into rapidly varying motor commands and slower changing higher-level goals. Similarly, different types of information tend to be processed by different levels of the system (\eg\ vision vs. proprioception).

The hierarchical formulation also gives rise to a modular structure that allows to selectively constrain, transfer and generalize certain aspects of the behavior, such as low-level skills or high-level goals. Exploiting the compositionality of probabilistic models, this modularity also facilitates the immediate reuse of some model components in a principled manner, such as the sharing of low-level skills across tasks. This allows imposing more flexible constraints and can reduce the number of model parameters leading to an increased statistical efficiency and faster learning.

The presence of latent variables introduces additional  complexities and we present a general  framework which addresses these 
in Appendix~\ref{appendix:general_framework}. We elaborate on a particular but still general hierarchically structured variant in the main text, and develop suitable efficient off-policy algorithms.
We provide empirical results on several tasks with physically simulated bodies and continuous action spaces and discrete grid worlds
which demonstrate the benefits of the general framework and especially the advantages of the hierarchical structure, compared to the unstructured policies used in \cite{galashov2018information}. 

\fi

\section{RL as probabilistic modelling}
\label{sec:prob_rl}

In this section, we briefly review how the KL-regularized objective can connect RL and probabilistic model learning.
We will denote states and actions at time $t$ respectively with $s_t$ and $a_t$. $r(s,a)$ is the instantaneous reward received in state $s$ when taking action $a$. We will refer to the history up to time $t$ as $x_t = (s_1, a_1, .., s_t)$ and the whole trajectory as $\traj=(s_1, a_1, s_2, a_2, \ldots)$.  The agent policy $\qq(a_t|x_t)$ denotes a distribution over next actions given history $x_t$, while $\pp(a_t|x_t)$ denotes a default or habitual policy.\footnote{
We generally work with history dependent policies since we will consider restricting access to state information from policies (for information asymmetry), which may render fully observed MDPs effectively partially observed.
}
The KL-regularized RL objective \cite{todorov2007linearly,kappen2012optimal,rawlik2012stochastic,schulman2017equivalence} takes the form:
\begin{align}
\textstyle
\mathcal{L}(\qq, \pp) & = \textstyle \EE_{\traj} \left[ 
\sum_{t\ge1}  \gamma^t r(s_t,a_t) 
      - \alpha \gamma^t \KL(a_t|x_t) \right]
\label{eq:objective:KL_regularized}
\end{align}
where $\gamma$ is discount factor and $\alpha$ controls the relative contributions of both terms.
$\EE_{\traj}[\cdot]$ is taken w.r.t.\ the  trajectory distribution given by  agent policy and system dynamics: $p(s_1) \prod_{t\ge1} \qq(a_t | x_t) p(s_{t+1} | s_t, a_t)$.
We use a convenient notation\footnote{In the following, $\KL(Y|X)$ always denotes $\EE_{\qq(Y|X)}[\log\frac{\qq(Y|X)}{\pp(Y|X)}]$ for arbitrary variables $X$ and $Y$.} for the KL divergence: $\KL(a_t|x_t)=\EE_{\qq(a_t|x_t)}[\log\frac{\qq(a_t|x_t)}{\pp(a_t|x_t)}]$.

When optimized with respect to $\qq$ the objective can be seen to trade off expected reward with closeness (in terms of KL) between trajectories produced by executing $\qq$ and  $\pp$ (Appendix~\ref{appendix:general_framework}). 
This is also evident from the optimal $\qq$ in eq.~\eqref{eq:objective:KL_regularized} 
\begin{align}
    \textstyle \qq^*(a_t|x_t) & \textstyle= \pp(a_t|x_t) \exp \frac{1}{\alpha} (Q^*(x_t, a_t)-V^*(x_t)) \label{eq:objective:policy}\hspace{-0.1cm}\\
    \textstyle  Q^*(x_t,a_t) &= \textstyle r(s_t, a_t) + \gamma \EE_{s_{t+1} | s_t,a_t}[V^*(x_{t+1})]\\
    \textstyle  V^*(x_t)& \textstyle =\alpha\log \int \pi_0(a_t|x_t) \exp \frac{1}{\alpha}Q^*(x_t, a_t) da_t,
\end{align}
where $Q^*(\cdot)$ and $V^*(\cdot)$ are optimal action value and value functions of eq.~\eqref{eq:objective:KL_regularized}; See \citep[e.g.][]{rawlik2012stochastic,fox2016taming,schulman2017equivalence,nachum2017bridging} for derivations.
We can thus think of $\qq$ as a specialization of $\pp$ that is obtained by tilting $\pp$ towards high-value actions (as measured by the action value $Q$). 

As discussed in \cite{galashov2018information} the default behavior thus bears resemblance to a trajectory prior in a particular probabilistic model. Several recent works have considered optimizing eq.~\eqref{eq:objective:KL_regularized} when $\pp$ is of a fixed and simple form. For instance, when $\pp$ is chosen to be uniform  entropy-regularized objective is recovered e.g.~\cite{ziebart2010modeling,fox2016taming,haarnoja2017reinforcement,schulman2017equivalence,hausman2018learning}.
More interestingly, in some cases $\pp$ can be used to inject detailed prior knowledge into the learning problem. In a transfer scenario $\pp$ can be a \emph{learned} object, and the KL term plays effectively the role of a shaping reward. 

$\qq$ and $\pp$ can also be co-optimized. In this case the relative parametric forms of $\pp$ and $\qq$ are of importance.
The optimal $\pp$ in eq.~\eqref{eq:objective:KL_regularized} is
\begin{align}
    \textstyle
    \pp^*(a_t | x_t) & = \arg \max_{\pp} \EE_{\qq(a | x_t)} [ \log \pp(a | x_t) ]
    \label{eq:objective:default_policy},
\end{align}
which maximizes only terms in eq.~\eqref{eq:objective:KL_regularized} depending on $\pp$.
Thus learning $\pp$ can be seen as supervised learning where $\pp$ is trained to match the history-conditional action sequences produced by $\qq$. 
It should be clear that $\pp=\qq$ is the optimal solution when $\pp$ has sufficient information and capacity, and in this scenario the regularizing effect of $\pp$ is lost. When the information or the capacity of $\pp$ is limited 
then $\pp$ will be forced to generalize the behavior of $\qq$.
%
For instance, \cite{teh2017distral} and \cite{galashov2018information} consider a multitask scenario in which $\qq$ is given task-identifying information, while $\pp$ is not. 
As a result, $\pp$ is forced to learn a marginal trajectory distribution over tasks, which can be considered as a common default behaviour, and this behaviour is then shareable across tasks to regularize $\pi$.
More generally, appropriate choices for the model classes of $\pp$ and $\qq$ will allow us to influence both the learning dynamics as well as the final solutions to eq.~\eqref{eq:objective:KL_regularized} and thus provide us with a 
means of injecting prior knowledge into the learning problem.

\begin{figure}
    \centering
    \includegraphics[height=2.0in]{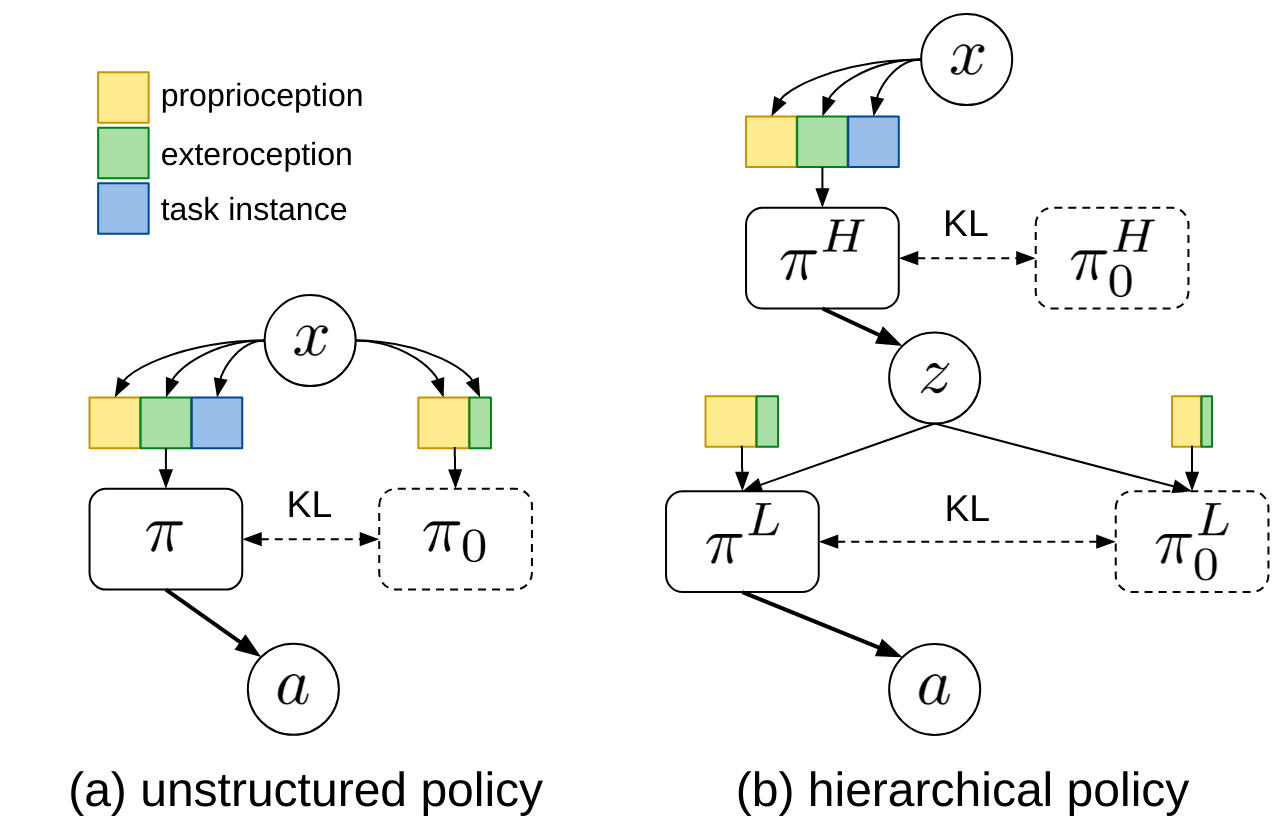}
    \vspace{-0.2cm}
    \caption{
        Diagram of the generic structure of the regularized KL-objective considered. (a) shows an unstructured policy, where information asymmetry (e.g.\ hiding task instance information) is exploited to induce a meaningful default policy $\pi_0$  \cite{galashov2018information}; (b) shows the scenario when we use structured policies composed from high-level $\pi^H$ and low-level $\pi^L$ policies that communicate through the latent action $z$. Note that now different forms of information asymmetry can be employed. See text for details. 
    }
    \vspace{-0.2cm}
    \label{fig:structured_policies}
\end{figure}

\section{Hierarchically structured policies}
\label{sec:hierarchy:KL}
\label{sec:hierarchy}

\cite{galashov2018information} consider simple unstructured models and focus on the information that $\pp$ has access to.
In this paper we focus on a complementary perspective. We explore how variations of the parametric forms of $\pp$ and $\qq$, via the introduction of latent variables, can give rise to richer and hierarchically structured models with different inductive biases and generalization properties. In this section we discuss a particular instantiation of this idea and discuss the general framework in Appendix~\ref{appendix:general_framework}.




We consider learning multi-level representations of behaviour.
Specifically, we consider a two level architecture in which the high-level decisions are concerned with task objectives but largely agnostic to details of actuation. The low-level control translates the high-level decisions into motor actions while being agnostic to task objectives.
Successful learning of such abstractions can be useful to exploit repetitive structures within or across tasks.
As two use cases we consider (a) multi-task control where different tasks require similar motor-skills; as well as (b) a scenario where we aim to solve similar tasks with different actuation systems.

We are interested in the role of the KL-regularized objective and the default policies for learning the desired multi-level abstractions.
In this view, the hierarchical structure and latent variables mirror the structure of the abstractions that we want to learn, and the structured default policy allows fine-grained control of its regularization effect in each level of abstractions.
Besides, the latent variables allow us to work with richer (e.g.\ non-Gaussian) distribution classes, allow us to model temporal correlations, e.g.\ in the high-level goals, and they give rise to modular structure that enables parameter sharing.

Conceptually policies are divided into high-level and low-level components which interact via auxiliary latent variables.  
For concreteness, but without loss of generality, let $z_t$ be a (continuous) latent variable for each time step $t$ (we discuss alternative choices such as latent variables that are sampled infrequently in Appendices~\ref{appendix:general_framework}~and~\ref{appendix:derivations}). The agent policy is extended as $\qq(a_t,z_t|x_t)= \qq^H(z_t|x_t)\qq^L(a_t|z_t,x_t)$ and likewise for the default policy $\pp$.  $z_t$ can be interpreted as a high-level or abstract action, taken according to the high-level (HL) controller $\qq^H$, and which is translated into low-level or motor action $a_t$ by the low-level (LL) controller $\qq^L$.  We extend the histories $x_t$ and trajectories $\traj$ to appropriately include $z_t$'s. Note that as we elaborate in the appendix this allows for temporally correlated $z_t$'s including the case where $z$ is only sampled once at the beginning of the episode.
As will be discussed in Section~\ref{sec:related_work}, structuring a policy into HL and LL controllers has been studied e.g.~\cite{heess2016learning,hausman2018learning,haarnoja2018latent,merel2018neural}, but the concept of a default policy has not been widely explored in this context. We discuss the differences between these works and ours in detail in Appendix \ref{appendix:general_framework}.

In case $z_t$'s can take on many values or are continuous, the objective \eqref{eq:objective:KL_regularized} becomes intractable as the marginal distributions $\qq(a_t|x_t)$ and $\pp(a_t|x_t)$ in the KL divergence cannot be computed in closed form. As discussed in more detail in  Appendix~\ref{appendix:general_framework} this problem can be addressed in different ways. For simplicity and concreteness we here assume that the latent variables in $\qq$ and $\pp$ have the same dimension and semantics.
We can then construct a lower bound for the objective by using the following upper bound for the  KL:
\begin{align}
\KL(a_t|x_t)
\le& \KL(z_t|x_t) + \EE_{\qq(z_t|x_t)}[\KL(a_t|z_t,x_t)],
\label{eq:kl_upper_bound}
\end{align}
which is tractably approximated using Monte Carlo sampling.
The derivation is in Appendix~\ref{appendix:sub:kl_upper_bound}. 
Note that:
\begin{align}
\KL(z_t|x_t) &= \KL\big(\pi^H(z_t|x_t)\|\pi^H_0(z_t|x_t)\big) \\
\KL(a_t|z_t,x_t) &= \KL\big(\pi^L(a_t|z_t,x_t)\|\pi^L_0(a_t|z_t, x_t)\big).
\end{align}
The resulting lower bound for the objective is:
\begin{equation}
\begin{split}
\textstyle
\mathcal{L}(\qq, \pp) & \textstyle \geq \textstyle \EE_{\traj}\Big[
\sum_{t\ge1} \gamma^t r(s_t,a_t)
    - \alpha\gamma^t\KL(z_t|x_t)\\
    &\hspace{1.1cm}\textstyle- \alpha\gamma^t\KL(a_t|z_t,x_t)\Big],
\label{eq:full_objective}
\end{split}
\end{equation}
where $\traj$ is a trajectory that appropriately includes $z_t$'s.
Full derivation including discount terms is in Appendix~\ref{appendix:sec:sec:discount_lower_bound_derivation}.
In this paper we consider eq.~\eqref{eq:full_objective} as a main objective function.






\subsection{Structuring the default policy}
Compared to e.g.\ \cite{galashov2018information} who mostly use conditional Gaussian distributions to model the default policy, the presence of latent variables may allow more flexible trajectory models that can, in principle, capture richer, non-Gaussian state-conditional distributions as a marginal $\pp(a_t|x_t)=\int_{z_t}\pp^H(z_t|x_t)\pp^L(a_t|z_t,x_t)d{z_t}$, and thus to approximate $\pp^*$ more accurately.\footnote{
Note that $\pp^*$ is itself obtained by marginalizing  over different contexts, e.g.~different task-specific policies, cf.\ eq.\ (\ref{eq:objective:default_policy})). Thus, even though the individual task-conditional policies may be well represented by conditional Gaussian distributions, their mixture may not be.
}
We can also model different marginal distribution based on the parameterization of $\pp^H(z_t|x_t)$, which could be a useful inductive bias for generalizing the learned trajectory distribution.

In this work, we consider the following choices of of HL default policy:
\textbf{Independent isotropic Gaussian.} $\pi^H_0(z_t | x_t) = \mathcal{N}(z_t | 0, 1)$, i.e.\ the HL default policy assumes the abstract actions to be context independent.
\textbf{AR(1) process.} $\pi^H_0(z_t | x_t) = \mathcal{N}(z_t | \alpha z_{t-1}, \sqrt{1-\alpha^2} )$, i.e.\ the HL default policy is a first-order auto-regressive process with a fixed parameter $0 \leq \alpha < 1$ chosen to  ensure a marginal distribution $\mathcal{N}(0,1)$. This allows for more structured temporal dependence among the abstract actions.
\textbf{Learned AR process.} Similar to the AR(1) process this default HL policy allows $z_t$ to depend on $z_{t-1}$ but now the high-level default policy is a Gaussian distribution with mean and variance that are learned functions  of $z_{t-1}$ with parameters $\phi$:
$\pi^H_0(z_t | x_t) = \mathcal{N}(z_t | \mu_\phi(z_{t-1}), \sigma^2_\phi(z_{t-1}) )$.
Note that the considered HL default policies are not conditioned on $x_t$. This is a form of information asymmetry for capturing task agnostic trajectory distribution as we will discuss in the following.

\paragraph{Regularizing via information asymmetries}
\label{sec:hierarchy:asymmetry}

As discussed in \cite{galashov2018information} restricting the information available to different policies is a powerful tool to force regularization and generalization. In our case we let this information asymmetry be reflected also in the separation between HL and LL controllers (see Figure \ref{fig:structured_policies}). Specifically we introduce a separation of concerns between $\qq^L$ and $\qq^H$ by providing full information only to $\qq^H$ while information provided to $\qq^L$ is limited. %
In our experiments we vary the information provided to $\qq^L$; it receives body-specific (proprioceptive) information as well as different amounts of environment-related (exteroceptive) information. The task is only known to $\qq^H$.
Hiding task specific information from the LL controller makes it easier to transfer across tasks. It forces $\qq^L$ to focus on learning task agnostic behaviour, and to rely on the abstract actions selected by $\qq^H$ to solve the task.
Similarly, we hide task specific information from both $\pp^L$ and $\pp^H$, which in turn makes the marginal default policy $\pp$ to be agnostic to the task (see above).
%
In the experiments we further consider transferring the HL controller across bodies, in situations where the abstract task is the same but the body changes. Here we additionally hide body-specific information from $\qq^H$, so that the HL controller is forced to learn body-agnostic behaviour.

\paragraph{Partial parameter sharing}
\label{sec:hierarchy:sharing}


An advantage of the hierarchical structure of the policies is that it enables several options for partial parameter sharing, which when used in the appropriate context can make learning more statistically efficient.
%
We explore the utility of sharing low-level controllers between agent and default policy, i.e.\ $\pi^L(a_t|z_t,x_t)=\pi^L_0(a_t|z_t,x_t)$, and study the associated trade-offs in different learning scenarios.
In the case of sharing, the new lower bound is:
\begin{equation}
\begin{split}
\textstyle
\mathcal{L}(\qq, \pp) & \textstyle \geq \textstyle \EE_{\traj}\left[
\sum_{t\ge1} \gamma^t r(s_t,a_t)
    - \alpha\gamma^t\KL(z_t|x_t) \right]\hspace{-0.1cm}
\label{eq:latent_lower_bound}
\end{split}
\end{equation}
This objective function is similar in spirit to current KL-regularized RL approaches discussed in Section \ref{sec:prob_rl}, except that the KL divergence is between policies defined on abstract actions $z_t$ as opposed to concrete actions $a_t$. The effect of this KL divergence is that it regularizes both the HL policies as well as the space of behaviours parameterised by the abstract actions.
%
This special case of our framework also reveals a connection to \cite{goyal2018transfer}, where eq.~\eqref{eq:latent_lower_bound} was motivated as an approximation of information botteneck for learning a goal conditioned policy.
Here, we provide a different perspective of eq.~\eqref{eq:latent_lower_bound} in terms of probabilistic modeling and structured policies as a special case of eq.~\eqref{eq:full_objective} (and more generally the framework in Appendix A), which suggests a broader range of algorithms that were not discussed in~\cite{goyal2018transfer}; see Section~\ref{sec:related_work} for detailed discussion.

\begin{algorithm}[tb]
   \caption{On-policy verison of our algorithm}
   \label{alg:actor_critic}
\begin{algorithmic}
   \STATE Flat (HL$+$LL) policy: $\pi_\theta(a_t|\epsilon_t,x_t)$, parameter $\theta$
   \STATE Reparameterized HL policy: $z_t=f^H_\theta(x_t, \epsilon_t)$
   \STATE Default policy: $\pi^H_{0,\phi}(z_t|x_t)$, $\pi^L_{0,\phi}(a_t|z_t,x_t)$, parameter $\phi$
   \STATE Q-function: $Q_\psi(a_t,z_t,x_t)$, parameter $\psi$
   \REPEAT
   \FOR{$t=0, K, 2K, ... T$}
   \STATE Rollout trajectory: $\tau_{t:t+K}=(s_t,a_t,r_t,...,r_{t+K})$
   \STATE Sample latent: $z_{t^\prime}=f^H(x_{t^\prime},\epsilon_{t^\prime})$, where $\epsilon_{t^\prime} \sim \rho(\epsilon)$
   \STATE Compute KL: $\hat{\KL}_{t^\prime}=\KL\left(z|x_{t^\prime}\right)+\KL\left(a|z_{t^\prime},x_{t^\prime}\right)$
   \STATE Bootstrap:$\hat{V}=\EE_{\pi}Q(a,z_{t+K},x_{t+K})-\alpha\hat{\KL}_{t+K}$
   \STATE Estimate Q target: $\hat{Q}_{t^\prime}=\sum_{i=t^\prime}^{t^\prime+K-1}(r_i-\alpha\hat{\KL}_i) + \hat{V}$
   \STATE Policy loss: $\hat{L}_\pi=\sum_{i=t}^{t+K-1}\EE_{\pi}Q(a,z_{i},x_{i})-\alpha\hat{\KL}_i$
   \STATE Q-value loss: $\hat{L}_Q=\sum_{i=t}^{t+K-1}\|\hat{Q}_i-Q(a,z_{i},x_{i})\|^2$
   \STATE Default policy loss: $\hat{L}_{\pi_0}=\sum_{i=t}^{t+K-1}\hat{\KL}_i$
   \STATE
   $\theta \leftarrow \theta+\beta_\pi\nabla_\theta\hat{L}_\pi$ \hspace{0.3cm}
   $\phi \leftarrow \phi+\beta_{\pi_0}\nabla_\phi\hat{L}_{\pi_0}$
   \STATE$\psi \leftarrow \psi-\beta_{Q}\nabla_\psi\hat{L}_{Q}$
   \ENDFOR
   \UNTIL
\end{algorithmic}
\end{algorithm}

\section{Algorithm}

We jointly optimize the default policy and the agent's policy, while the agent's policy is regularized by a \textit{target default policy}, which is periodically updated to a new default policy.
The objective for the default policies is similar to distillation 
\cite{parisotto2015actor,rusu2015policy} or supervised learning, where agent's policies define the data distribution. Note that due to the particular way we lower bound the KL (see eq.~\eqref{eq:full_objective}) the supervised step remains unproblematic despite the presence of the latent variables in $\pp$.

To efficiently optimize the hierarchical policy with latent variables, we develop off-policy learning algorithm based on SVG(0)~\cite{heess2015learning} with experience replay.
We follow a strategy similar to \cite{heess2016learning} and reparameterize $z_t\sim \pi^H(z_t|x_t)$ as $z_t = f^H(x_t, \epsilon_t)$, where 
$\epsilon_t \sim \rho(\epsilon_t)$ is a fixed noise distribution and
$f^H(\cdot)$ is a deterministic function. 
In practice this means that the hierarchical policy can be treated as a flat policy $\pi(a_t|\epsilon_t,x_t)=\pi^L(a_t|f^H(x_t, \epsilon_t), x_t)$.
We reparameterize action distribution of the flat policy $\pi(a_t|\epsilon_t,x_t)$ and optimize it by backpropagating the gradient from an action value function $Q(a_t, z_t, x_t)$.
Note that the action value function is conditioned on $z_t$ because it will be contained in history from the next time step and affect the policy accordingly.
The action value function is optimized to match a target action value estimated by Retrace~\cite{munos2016safe}, which provides low variance estimate of action value from K-step windows of off-policy trajectories.
Note that the off-policy learning with hierarchical policy requires extra care in terms of handling latent variables in action value function and applying Retrace.
We describe these details in Appendix~\ref{appendix:algorithm}.
For illustration, Algorithm~\ref{alg:actor_critic} provides the pseudo-code for a simple on-policy version of our algorithm.
We implement our algorithm in a distributed setup similar to \cite{riedmiller2018learning} where multiple actors are used to collect trajectories and a single learner is used to optimize model parameters.
Similarly to other KL-regularized RL approaches e.g.~\cite{teh2017distral,galashov2018information}, we additionally regularize the entropy of $\qq^L$ to encourage exploration.
More details about the learning algorithms are in Appendix~\ref{appendix:algorithm}.

\begin{figure*}[t]
    \centering
    \includegraphics[width=.9\linewidth]{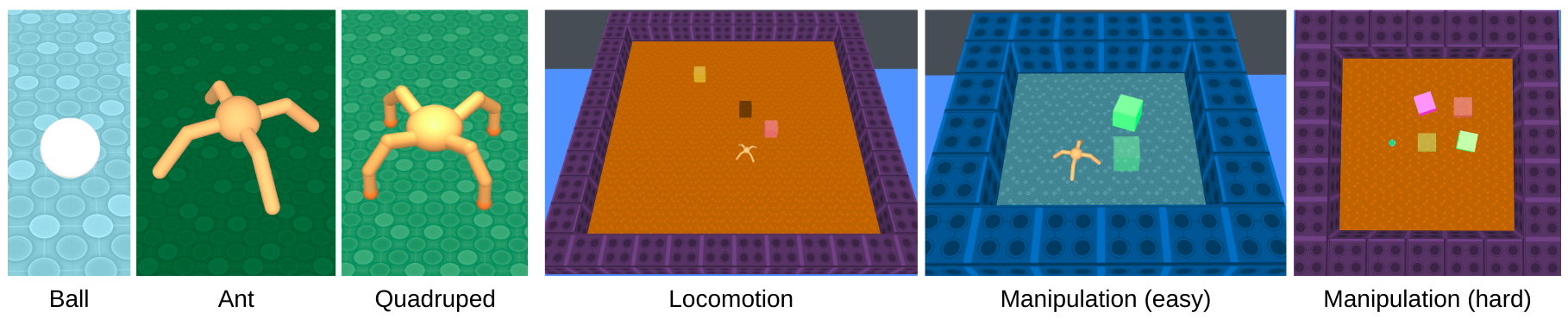}
    \vspace{-0.4cm}
    \caption{
        \textbf{Bodies and tasks for experiments.} \textbf{Left}: All considered bodies. \textbf{Right}: Example tasks. 
    }
    \vspace{-0.2cm}
    \label{fig:continuous_environment}
\end{figure*}

\begin{figure*}[t]
    \centering
    \includegraphics[width=0.9\linewidth]{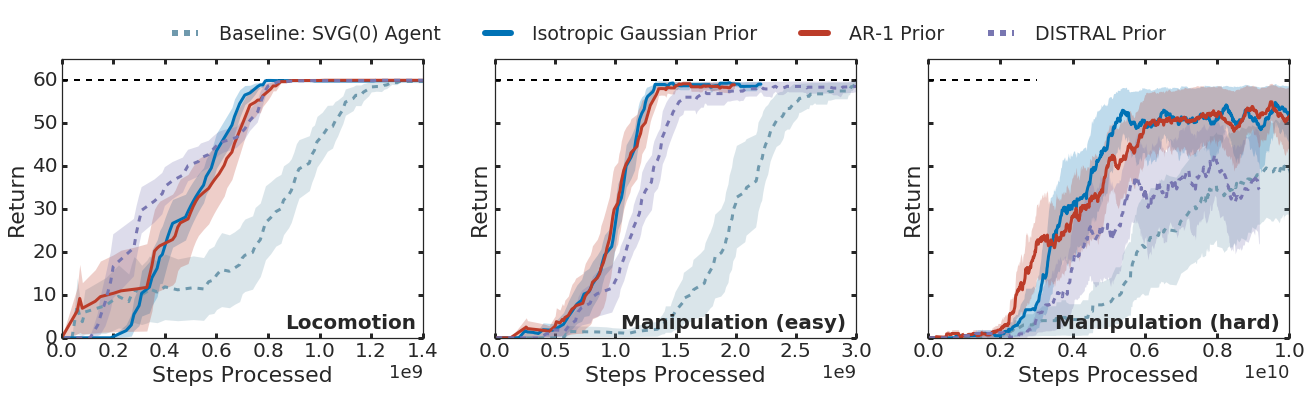}
    \vspace{-0.3cm}
    \caption{
        \textbf{Learning with structured policies.} \textbf{Left}: Locomotion with the Ant. \textbf{Center}: Manipulation (easy) with the Ant. \textbf{Right}: Manipulation (hard) with the Ball. The proposed models with hierarchy are denoted with the type of HL default policy: Isotropic Gaussian, AR-1, AR-Learned.
    }
    \vspace{-0.4cm}
    \label{fig:scratch_continuous}
\end{figure*}

\section{Related Work}
\label{sec:related_work}

Entropy regularized reinforcement learning (RL), also known as maximum entropy RL~\cite{ziebart2010modeling,kappen2012optimal,toussaint2009robot} is a special case of KL regularized RL.
This framework connects probabilistic inference and sequential decision making problems.
Recently, this idea has been adapted to deep reinforcement learning~\cite{fox2016taming,schulman2017equivalence,nachum2017bridging,haarnoja2017reinforcement,hausman2018learning,haarnoja2018soft}.
Another instance of KL regularized RL includes trust region based methods~\cite{schulman2015trust,schulman2017proximal,wang2017sample,abdolmaleki2018maximum}. They use KL divergence between new policy and old policy as a trust region constraints for conservative policy update.

Introducing a parameterized default policy provides a convenient way to transfer knowledge or regularize the policy.
Schmitt et al.~\cite{schmitt2018kickstarting} use a pretrained policy as the default policy;
other works jointly learn the policy and default policy to capture reusable behaviour from experience~
\cite{teh2017distral,czarnecki2018mix,galashov2018information,grau-moya2018soft}.
To retain the role of default policy as a regularizer, it has been explored to restrict its input~\cite{galashov2018information,grau-moya2018soft}, parameteric form~\cite{czarnecki2018mix} or to share it across different contexts~\cite{teh2017distral,ghosh2018divideandconquer}.

Another closely related regularization for RL is using information bottleneck~\cite{tishby2011information,still2012information,rubin2012trading,ortega2013thermodynamics,tiomkin2017unified}. Galashov et al.~\cite{galashov2018information} discussed the relation between information bottleneck and KL regularized RL.
Strouse et al.~\cite{strouse2018learning} learn to hide or reveal information for future use in multi-agent cooperation or competition.
Goyal et al.~\cite{goyal2018transfer} consider identifying bottleneck states based on eq.~\eqref{eq:latent_lower_bound} and using them for exploration during transfer.
However, unlike \cite{goyal2018transfer} motivating the objective simply as an approximation to an information bottleneck, we provide a new perspective that considers structuring policies with latent variables as a way of injecting inductive bias in the KL-regularized RL framework.
This new perspective motivates the eq.~\eqref{eq:latent_lower_bound} as one instance of a broader family of algorithms that were not discussed in~\cite{goyal2018transfer}, and we demonstrate how different instances of the proposed family of algorithms can be useful in different learning scenarios.

The hierarchical RL literature ~\cite{dayan1993feudal,parr1998reinforcement,sutton1999between} has studied hierarchy extensively as a means to introduce  inductive bias.
Among various ways~\cite{sutton1999between,bacon2017option,vezhnevets2017feudal,nachum2018data,nachum2018nearoptimal,xie2018transferring,lee2018composing}, our approach resembles~\cite{heess2016learning,hausman2018learning,haarnoja2018latent,merel2018neural}, in that a HL controller modulates a LL controller through a continuous channel.
For learning the LL controller, imitation learning~\cite{fox2017multi,krishnan2017discovery,merel2018neural}, unsupervised learning~\cite{gregor2016variational,eysenbach2018diversity} and meta learning~\cite{frans2018meta} have been employed.
Similar to our approach, ~\cite{heess2016learning,florensa2017stochastic,hausman2018learning} use a pretraining task to learn a reusable LL controller. However, the concept of a default policy has not been widely explored in this context.


\section{Experiments}
\label{sec:experiments}

We evaluate our method in several environments with continuous state and action spaces.
We consider a set of structured, sparse reward tasks that can be executed by multiple bodies with different degrees of freedom.
The tasks and bodies are illustrated in Figure~\ref{fig:continuous_environment}.
%

We consider task distributions that are designed such that their solutions exhibit significant overlap in trajectory space so that transfer can reasonably be expected. They are further designed to contain instances of
variable difficulty and hence provide a natural curriculum.
Our tasks are as follows. \textbf{Locomotion}: reaching a specific target among 3 locations. \textbf{Locomotion with gap}: reaching the end of a corridor with a gap of variable length. Solving the task requires being able to jump over the gap.
\textbf{Manipulation}: moving one of N boxes to one of K targets as indicated by the environment. (easy): N=1, K=1, (mid): N=1, K=3, (hard): N=2, N=2.
\textbf{Manipulation (heavy)}: manipulation (easy) with heavier box.
\textbf{Manipulation (gather)}: moving two boxes such that they are in contact with each other.
\textbf{Combined task (And / Or)}: moving a box to one target (And / Or) go to the other target in a single episode.
See Appendix~\ref{appendix:environments} for exact configurations.

We use three simulated bodies: Ball, Ant, and Quadruped.
Ball and Ant have been used in~\cite{heess2017emergence,xie2018transferring,galashov2018information}; we introduce the Quadruped as a variant of the Ant.
The \textbf{Ball} has 2 actuators for moving forward or backward and turning left or right.
The \textbf{Ant} has 8 actuators for moving its legs to walk and to interact with objects.
The \textbf{Quadruped} is similar to the Ant, but with 12 actuators.
Each body is characterized by a different set of proprioceptive features.
Further details of the bodies are in Appendix~\ref{appendix:environments}.

Throughout the experiments, we use 32 actors to collect trajectories and a single learner to optimize the model.
We plot average episode return with respect to the number of steps processed by the learner.
Note that the number of steps is different from the number of agent's interaction with environment, because the collected trajectories are processed multiple times by a centralized learner to update model parameters.
Hyperparameters, including KL cost and action entropy regularization cost, are optimized on a per-task basis.
Details are provided in Appendices~\ref{appendix:experimental_settings} and~\ref{appendix:hyper_parameters}.

\subsection{Learning with structured policies}

We study how the KL-regularized objective with different structures and parameterizations discussed in Section~\ref{sec:hierarchy} affects learning.
Our baselines are SVG-0 with entropy regularization and an unstructured KL regularized policy similar to \cite{galashov2018information,teh2017distral} (DISTRAL prior).
As described in Section~\ref{sec:hierarchy} we employ hierarchical structure with shared LL components (Shared LL) and separate LL components (Separate LL).
Unless otherwise stated, we use Shared LL as our default hierarchical model.
The HL controller receives full information while the LL controller (and hence the default policy) receives proprioceptive information plus the positions of the box(es) 
as indicated.
The same information asymmetry is applied to the DISTRAL prior i.e. the default policy receives proprioception plus box positions as inputs.
We explore multiple HL default policies: Isotropic Gaussian, AR(1) process, and learned AR prior.

Figure~\ref{fig:scratch_continuous} illustrates the results of the experiment.
Our main finding is that the KL regularized objective significantly speeds up learning, and that the hierarchical structure does as well or better than the flat, DISTRAL formulation. The gap increases for more challenging tasks (\eg~Manipulation (hard)).



\section{Transfer Learning}
\label{sec:transfer}

\begin{figure}[t]
    \centering
    \includegraphics[width=0.95\linewidth]{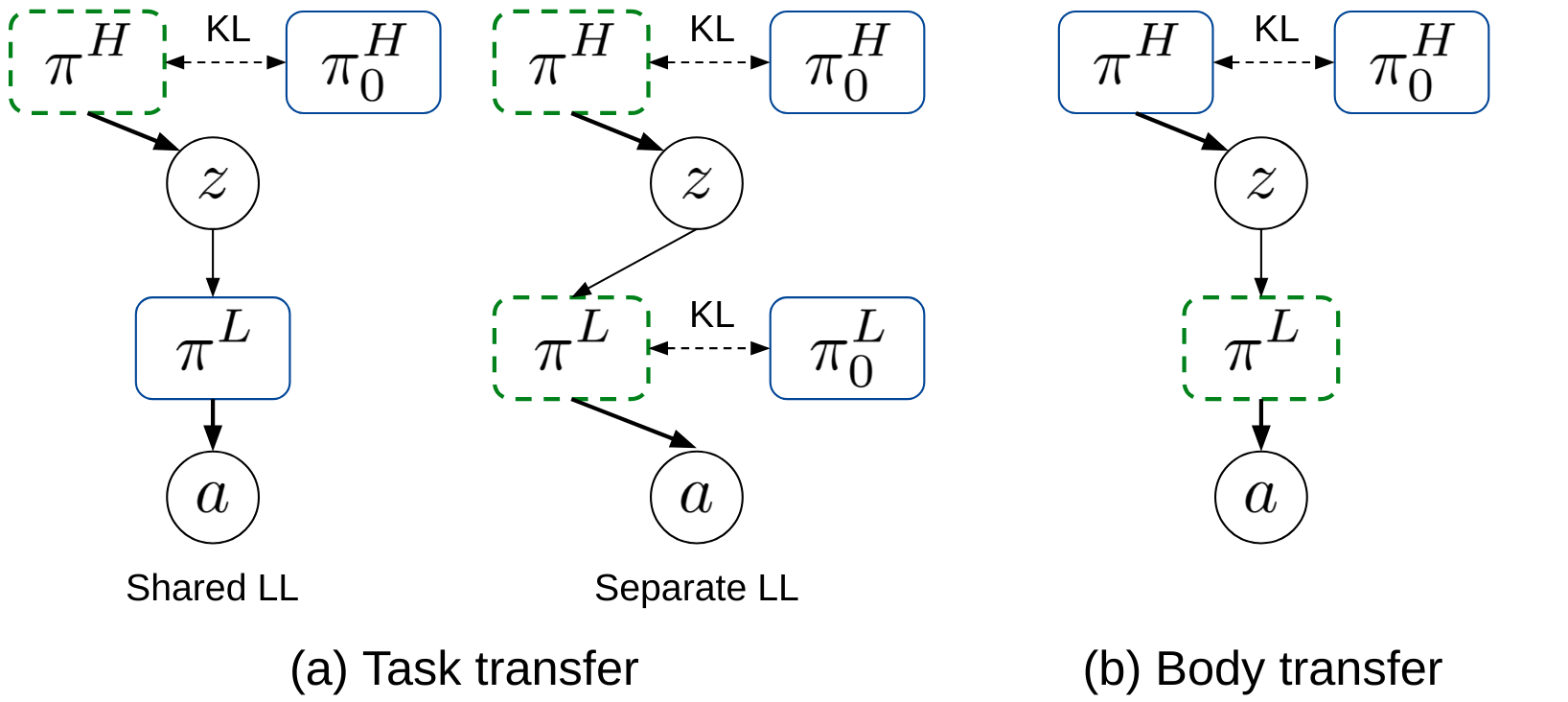}
    \vspace{-0.2cm}
    \caption{
        \textbf{Transfer learning scenarios.} The blue boxes denote modules that are transferred and fixed, and the green dotted boxes denote modules learned from scratch.
    }
    \label{fig:transfer_modular}
\end{figure}

The hierarchical structure introduces a modularity of the policy and default policy, which can be utilized for transfer learning.
We consider two transfer scenarios (see Figure~\ref{fig:transfer_modular}): 1) task transfer where we reuse the learned default policy to solve novel tasks, and 2) body transfer, where reusing the body agnostic HL policy and default policy transfers the goal directed behaviour to another body.

\subsection{Task transfer}

We consider transfer between task distributions whose solutions exhibit significant shared structure, e.g.\ because solution trajectories can be produced by a common set of skills or repetitive behaviour.
If the default policy can capture and transfer this reusable structure it will facilitate learning similar tasks. Transfer then involves specializing the default behavior to the needs of the target task (e.g.~by directing locomotion towards a goal).

For task transfer, we reuse pretrained goal agnostic components, including $\pp^H$ and $\pp^L$. We learn a new $\qq^H$ and either set the LL policy $\qq^L$ identical to the pretrained LL default policy $\pp^L$ (Shared LL), or allow $\qq^L$ to diverge from $\pp^L$ (Separate LL).
In the case of Shared LL, similarly e.g.\ to \cite{heess2016learning,hausman2018learning}, the new HL policy $\qq^H$ learns to control the LL policy $\qq^L$ in the latent space. Unlike in previous work, however, we regularize $\qq^H$ with $\pp^H$.
Transferring the learned default policy to a new task is similar to \cite{galashov2018information}, but our approach is different in that we exploit the modular structure of $\pp$ and $\qq$; It allows to re-use the pretrained $\qq^L$ (Shared LL) or to initialize the new LL policy with the pretrained parameters (Separate LL with weight initialization).

We consider two baselines: (a) the Shared LL model learned from scratch (Hierarchical Agent); (b) a DISTRAL prior, i.e.~we transfer a pretrained unstructured default policy to regularize the policy for the target task.
The first baseline allows us to assess the benefit of transfer; the second baseline provides an indication whether the hierarchical policy structure is beneficial.
Additionally, we compare different types of HL default policies. Specifics of the experiments including the information provided to HL and LL are provided in Appendix~\ref{appendix:environments}.

\begin{figure}[t]
    \centering
    \vspace{-0.2cm}
    \hspace{-0.3cm}\includegraphics[width=1.0\linewidth]{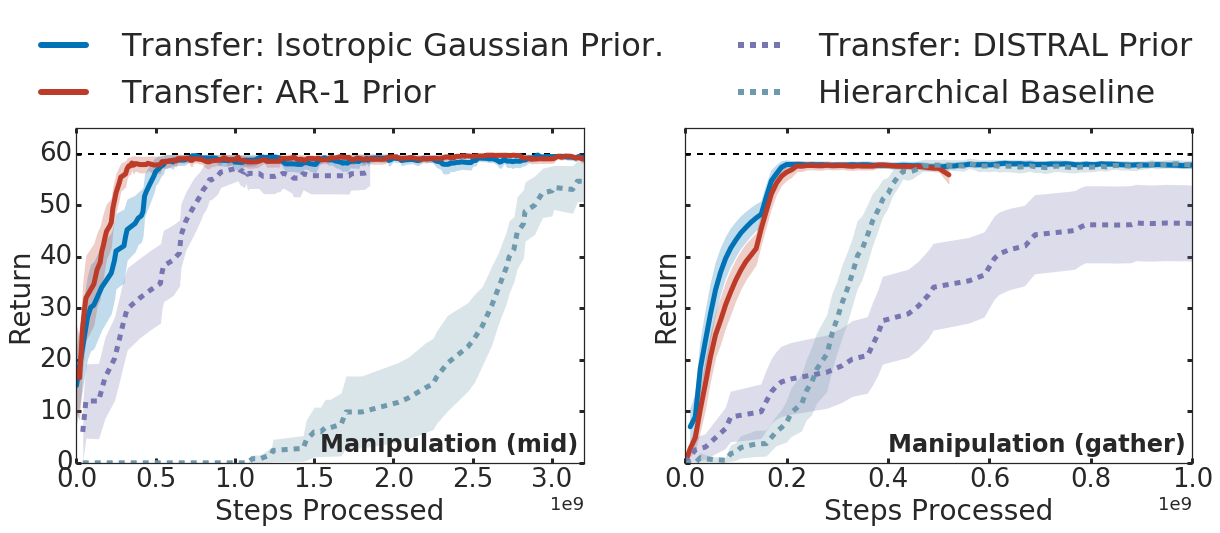}
    \vspace{-0.3cm}
    \caption{
        \textbf{Transfer learning with structured policies.} \textbf{Left}: From Manipulation (easy) to Manipulation (mid) with Ant. \textbf{Right}: From Manipulation (hard) to Manipulation (gather) with Ball.
    }
    \label{fig:task_transfer_continuous_summary}
    \vspace{-0.2cm}
\end{figure}

\begin{figure}[t]
    \centering
    \hspace{-0.3cm}\includegraphics[width=1.0\linewidth]{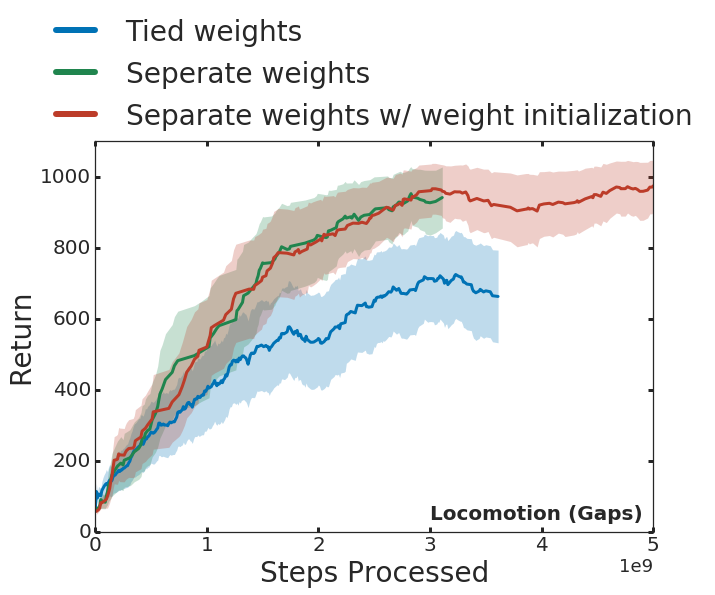}
    \vspace{-0.3cm}
    \caption{
        \textbf{Effect of partial parameter sharing.} From Locomotion (easy) to Locomotion (Gap) with Ant, AR-1.
    }
    \label{fig:task_transfer_seprate_ll}
    \vspace{-0.3cm}
\end{figure}

Figure~\ref{fig:task_transfer_continuous_summary} shows the results with Shared LL.
Transferring the pretrained default policy can bring clear benefits for related tasks.
The hierarchical structure, which facilitates partial parameter sharing and transferring flexible non-Gaussian default policy, performs better than the DISTRAL prior regardless of type of HL default policy.
To assess the benefit of the more flexible trajectory model we measure the KL divergence between $\qq$ and $\pp$ in Manipulation (mid) for the DISTRAL prior and the hierarchical latent variable model with Gaussian prior. Values of 11.35 and 2.64 respectively are consistent with the idea that the latent variable default policy provides a better prior over the trajectory distribution and thus allows for a more favorable trade-off between KL and task reward.
Figure~\ref{fig:task_transfer_seprate_ll} illustrates the benefits of the flexibility afforded by the full model in eq.\ \eqref{eq:full_objective} compared to eq.\ \eqref{eq:latent_lower_bound}.
In the presented transfer scenarios the adaptation of low level skills is required to learn the target task. Specifically, the transfer task requires the acquisition of a new skill (jumping) in order to consistently solve the task.
Allowing the $\qq^L$ to diverge from $\pp^L$ as in eq.\ (\ref{eq:full_objective}) turns out to be critical in these scenarios.

\subsection{Body transfer}

We study whether our architecture can facilitate transfer of high-level, goal-directed behavior between bodies with different actuation systems.
To this end we reuse the pretrained body-agnostic components, HL policy $\pi^H$ and the default policy $\pi^H_0$. We learn a new body-specific LL policy $\pi^L$, which is assumed to be shared with LL default policy $\pp^L$ (see Figure~\ref{fig:transfer_modular}b).
The transferred HL policies provide goal-specific behaviour actuated on the latent space, which can then be instantiated by learning a new LL policy.
During transfer, we optimize eq.~\eqref{eq:latent_lower_bound} to exploit KL between the HL policies as a dense reward signal to guide learning new LL policy.
Here, using the KL between the HL policies during transfer looks superficially similar to~\cite{goyal2018transfer}, but their motivation and method are different. They rely on the \emph{positive KL} as an exploration bonus, trying to get the agent to explore the new state. We rely on \emph{negative KL}, encouraging the new policy to be close to the default policy.
We discuss further differences with~\cite{goyal2018transfer} in Section~\ref{sec:related_work}.

\begin{figure}[t!]
    \centering
    \includegraphics[width=0.9\linewidth]{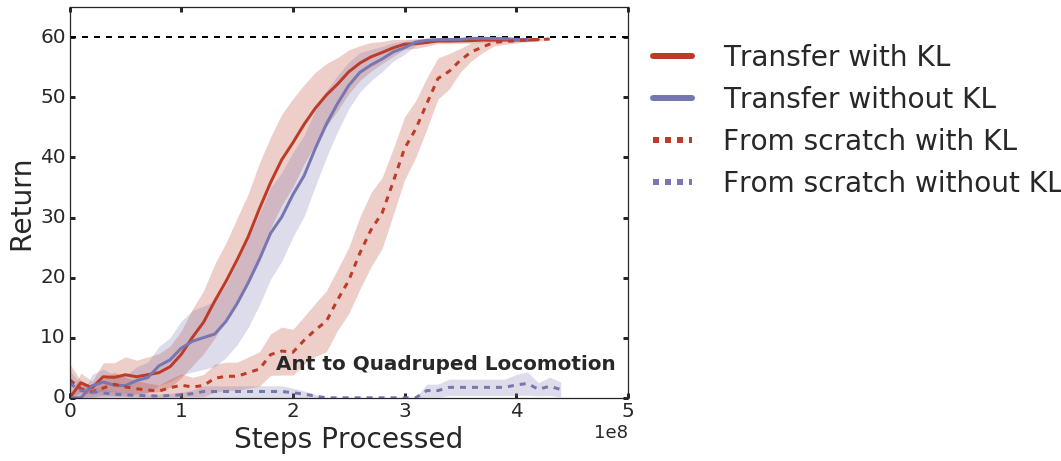}
    \vspace{-0.3cm}
    \caption{
        \textbf{Body transfer from vision, AR-1 Prior.} Ant to Quadruped, Locomotion. Task information is provided by egocentric vision.
    }
    \vspace{-0.1cm}
    \label{fig:body_transfer_vision}
\end{figure}

Figure~\ref{fig:body_transfer_vision} illustrates result for transferring from Ant to Quadruped in Locomotion task with egocentric vision as observation input.
We compare our approach with baseliness learning the hierarchical policy from scratch and analyze the effects of the KL regularization.
It shows that transferring the HL component and using KL regularization works best in this setting.
We observed that this result is consistent in multiple settings with different tasks and bodies both in continuous control and discrete grid world environments. These additional experiments can be found in Appendix~\ref{appendix:sub:additional_body_transfer}.
\section{Discussion}

In this work we have studied the benefit of learned probabilistic models of default behaviors for multi-task and transfer scenarios in RL. In particular, we have outlined a generic framework for hierarchically structured models together with suitable off-policy RL algorithms. The hierarchical model structure allows to model richer distributions, can give rise to modular network structure and generally allows to express a broad range of inductive biases with different generalization properties.
Here, we have studied a particular model variant and shown its empirical advantages. Looking forward we believe that as the importance of multitask and lifelong learning scenarios will grow, frameworks such as ours, that provide flexible mechanisms for introducing prior knowledge and reusing previous learned solutions will provide a fertile ground for advancing RL algorithms.

\bibliography{icml2019}
\bibliographystyle{icml2019}

\clearpage

\twocolumn[
\icmltitle{Appendix}
]

\appendix

\section{A general framework for RL as probabilistic modelling}
\label{appendix:general_framework}

In Sections~\ref{sec:prob_rl} and~\ref{sec:hierarchy:KL} of the main text we have introduced the KL-regularized objective and explored a particular formulation that uses latent variables in the default policy and policy (Section~\ref{sec:hierarchy:KL} and experiments). The particular choice in Section~\ref{sec:hierarchy:KL} arises as a special case of a more general framework which we here outline briefly. 

For both the default policy and for agent policy we can consider general directed latent variable models of the following form
\begin{align}
\pp(\tau) &= \textstyle \int \pp(\tau| y) \pp(y) d y,\\
\qq(\tau) &= \textstyle \int \qq(\tau | z) \qq(z) d z
\end{align}
where both $y$ and $z$ can be time varying, e.g.\ $y=(y_1, \dots y_T)$, and can be causally dependent on the trajectory prefix $x_t$, e.g.\ $y_t \sim p(\cdot | x_t)$ (and equivalently for $z$). The latent variables can further be continuous or discrete, and $y_t$ or $z_t$ can exhibit further structure (and thus include e.g.~binary variables that model option termination). 
The general form of the objective presented in the main text
\begin{align}
\textstyle
\mathcal{L}(\qq, \pp) & = \textstyle \EE_{\traj} \left[ 
\sum_{t\ge1}  \gamma^t r(s_t,a_t) 
      - \alpha \gamma^t \KL(a_t|x_t) \right],
\nonumber 
\end{align}
 remains valid regardless of the particular form of $\pp$ and $\qq$.
This form can be convenient when $\pp(a_t | x_t)$ and $\qq(a_t | x_t)$ are tractable (e.g.\ when $z$ or $y$ have a small number of discrete states or decompose conveniently over time, e.g.~as in \cite{fox2017multi,krishnan2017discovery}).


In general, however, latent variables in $\pp$ and $\qq$ may 
introduce the need for additional approximations. In this case different models and algorithms can be instantiated based on a) the particular approximation chosen there, as well as b) choices for sharing of components between $\pp$ and $\qq$.
A possible starting point when $\pp$ contains latent variables is the following lower bound to $\mathcal{L}$:
\begin{align}
\mathcal{L} 
&= \textstyle \EE_\qq[ \sum_t r(s_t,a_t)] - 
\KL[ \qq(\tau) || \pp(\tau) ]\\
&\geq \textstyle \EE_\qq \left [ \sum_t r(s_t,a_t) + \EE_f \left [ \log\frac{\pp(\tau,y)}{f(y|\tau)} \right ] \right] + \Ent [ \qq(\tau) ]\\
&= \textstyle \EE_\qq\left [ \sum_t r(s_t,a_t) + \EE_f \left [ \log \pp(\tau|y) \right ] \right. \nonumber\\
&\hspace{1.8cm}-\left. \KL [ f(y|\tau) || \pp(y) ] \right] + \Ent [ \qq(\tau) ]. \label{eq:objective:prior_ELBO}
\end{align}
If $y_t$ are discrete and take on a small number of values we can compute $f(y|\tau)$ exactly (e.g.~using the forward-backward algorithm as in \cite{fox2017multi,krishnan2017discovery}); in other cases we can learn a parameterized approximation to the true posterior or can conceivably apply mixed inference schemes \citep[e.g.][]{johnson2016composing}.

Latent variables in the policy $\qq$ can require an alternative approximation discussed e.g.\ in \cite{hausman2018learning}:
\begin{equation}
\begin{split}
    \mathcal{L} 
\geq \textstyle \EE_\qq\big[ &\textstyle\sum_t r(s_t,a_t) +  \log \pp(\tau) + \log g(z | \tau)  \\
&\textstyle+  \Ent [ \qq(\tau| z) ]   \big] + \Ent [ q(Z)], 
\label{eq:objective:KL_prior_posterior}
\end{split}
\end{equation}
where $g$ is a learned approximation to the true posterior $\qq(z|\tau)$. 
(But see e.g.\ \cite{haarnoja2018latent} who consider a parametric form for policies with latent variables for which the entropy term can be computed analytically and no approximation is needed.)
This formulation bears interesting similarities with diversity inducing regularization schemes based on mutual information \citep[e.g.][]{gregor2016variational,florensa2017stochastic} but arises here as an approximation to trajectory entropy. This formulation also has interesting connections to auxiliary variable formulations in the approximate inference literature \cite{salimans2014bridging,agakov2004an}.  

When both $\pp$ and $\qq$ contain latent variables eqs. (\ref{eq:objective:prior_ELBO},\ref{eq:objective:KL_prior_posterior}) can be combined. The model described in Section~\ref{sec:hierarchy:KL} in the main text then arises when the latent variable is ``shared'' between $\pp$ and $\qq$ and we effectively use the policy itself as the inference network for $\pp$: $f(y| \tau) = \prod_t \qq(y_t | x_t)$. In this case the objective simplifies to
\begin{align}
\textstyle \mathcal{L} 
&\geq \textstyle \EE_\qq\left [ \sum_t r(s_t,a_t) +  \log \frac{\pp(\tau|z)\pp(z)}{\qq(\tau|z)\qq(z)} \right ].
\end{align}
When we further set $\pp(\tau|z) = \qq(\tau|z)$ we recover the model discussed in the main text of the paper.

As a proof-of-concept for a model without a shared latent space, with latent variables in $\pi_0$ but not $\pi$, we consider a simple humanoid with 28 degrees of freedom and 21 actuators and consider two different tasks: 1) a dense-reward walking task, in which the agent has to move forward, backward, left, or right at a fixed speed. The direction is randomly sampled at the beginning of an episode and changed to a different direction half-way through the episode and 2) a sparse reward go-to-target task, in which the agent has to move to a target whose location is supplied to the agent as a feature vector similar to those considered in \cite{galashov2018information}.

\begin{figure*}
    \centering
    \includegraphics[width=0.95\linewidth]{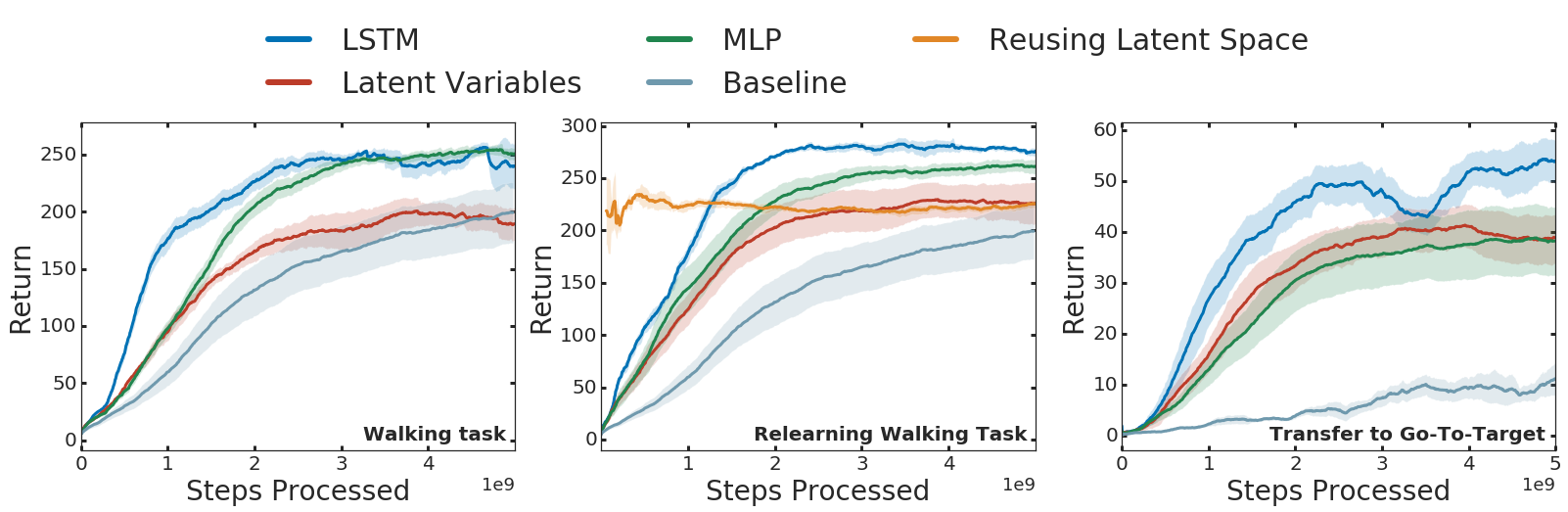}
    \caption{
        \textbf{Results with a latent variable prior.} \textbf{Left}: Walking task with the simple humanoid \textbf{Center}: Relearning the walking task with fixed priors. \textbf{Right}: Transfer to a go-to-target task.
    }
    \label{fig:latent_variables_in_prior}
\end{figure*}

Figure \ref{fig:latent_variables_in_prior} shows some exploratory results. In a first experiment we compare different prior architectures on the directional walking task. We let the prior marginalize over task condition. We include a feed-forward network, an LSTM, and a latent variable model with one latent variable per time step in the comparison. For the latent variable model we chose an inference network $f(z_t|z_{t-1}, \tau)$ so that eq.\ \eqref{eq:objective:prior_ELBO} decomposes over time. All priors studied in this comparison gave a modest speed-up in learning. While the latent variable prior works well, it does not work as well as the LSTM and MLP priors in this setup.
In a first set of transfer experiments, we used the learned priors to learn the walking task again. Again, the learned priors led to a modest speed-up relative to learning from scratch. 

We also experimented with parameter sharing for transfer as in the main text. We can freeze the conditional distribution $\pp(a|s, z)$ and learn a new policy $\qq(z|s)$, effectively using the learned latent space as an action space. 
In a second set of experiments, we study how well a prior learned on the walking task can transfer to the sparse go-to-target task. Here all learned priors led to a significant speed up relative to learning from scratch. Small return differences aside, all three different priors considered here solved the task with clear goal directed movements. On the other hand, the baseline only learned to go to very close-by targets. Reusing the latent space did not work well on this task. We speculate that the required turns are not easy to represent in the latent space resulting from the walking task.

\section{Algorithm}
\label{appendix:algorithm}

We develop efficient off-policy learning algorithms for optimizing hierarchically structured model with default policies.
Our off-policy learning algorithms are adapted from different existing algorithms based on the environments.
Specifically, the algorithm for continuous control environments is based on SVG(0)~\cite{heess2015learning} augmented with experience replay, and the algorithm for discrete action space environments is based on IMPALA~\cite{espeholt2018impala}, which is a framework for off-policy actor critic algorithm.
Note that the introduction of latent variables and default policies introduces additional challenges for off-policy learning algorithms in terms of gradient estimation of hierarchical policy, parameterization and estimation of value function, off-policy correction of the target value estimation, and appropriate incorporation of KL terms in the algorithm.
We describe how we address such challenges in this section.
Unless otherwise mentioned, we follow notations from the main paper.

\subsection{Reparameterized latent for hierarchical policy}

To estimate gradient for the hierarchical policy, we follow a strategy similar to \cite{heess2016learning} and reparameterize  $z_t\sim \pi^H(z_t|x_t)$ as $z_t = f^H(x_t, \epsilon_t)$, where $\epsilon_t \sim \rho(\epsilon_t)$ is a fixed distribution.
The $f^H(\cdot)$ is a deterministic function that outputs distribution parameters.
In practice this means that the hierarchical policy can be treated as a flat policy $\pi(a_t|\epsilon_t,x_t)=\pi^L(a_t|f^H(x_t, \epsilon_t), x_t)$. 
We exploit the reparameterized flat policy to employ existing distributed learning algorithm with minimal modification.

\subsection{Continuous control}
\label{sec:sec:alg_continuous}

In continuous control experiments, we employ distributed version of
SVG(0)~\cite{heess2015learning} augmented with experience replay and off-policy correction algorithm called Retrace~\cite{munos2016safe}.
In the distributed setup, behaviour policies in multiple actors are used to collect off-policy trajectories and a single learner is
used to optimize model parameters
The SVG(0) reparameterize a policy $p(a|s)$ and optimize it by backpropagating gradient from a learned action value function $Q(a,s)$ through a sampled action $a$.

To employ this algorithm, we reparameterize action from flat policy $a_t\sim \pi_\theta(a_t|\epsilon_t,x_t)$ with parameter $\theta$ as $a_t=h_\theta(\epsilon_t,x_t,\xi_t)$, where $\xi_t\sim\rho(\xi_t)$ is a fixed distribution, and $h_\theta(\epsilon_t,x_t,\xi_t)$ is a deterministic function outputting a sample from the distribution $\pi_\theta(a_t|\epsilon_t, x_t)$.
We also introduce the action value function $Q(a_t,z_t,x_t)$. Unlike policies without hierarchy, we estimate the action value depending on the sampled action $z_t$ as well, so that it could capture the future returns depending on $z_t$.
Given the flat policy and the action value function, SVG(0)~\cite{heess2015learning} suggests to use following gradient estimate
\begin{equation}
\begin{split}
&\nabla_{\theta}\EE_{\pi_\theta(a|\epsilon_t,x_t)}Q(a,z_t,x_t)\\
&=\nabla_{\theta}\EE_{\rho(\xi)}Q(h_\theta(\epsilon_t,x_t,\xi),z_t,x_t)\\
&=\EE_{\rho(\xi)}\frac{\partial Q}{\partial h}\frac{\partial h}{\partial \theta} \approx \frac{1}{M}\sum_{i=1}^{M}\frac{\partial Q}{\partial h}\frac{\partial h}{\partial \theta}\bigg|_{\xi=\xi_i},
\end{split}
\end{equation}
which facilitates using backpropagation.
Note that policy parameter $\theta$ could be learned through $z_t$ as well, but we decide not to because it tends to make learning unstable.

To learn action value function $Q(a_t,z_t,x_t)$ and learn policy, we use off-policy trajectories from experience replay.
We use Retrace~\cite{munos2016safe} to estimate the action values from off-policy trajectories.
The main idea behind Retrace is to use importance weighting to correct for the difference between the behavior policy $\mu$ and the online policy $\pi$, while cutting the importance weights to reduce variance.
Specifically, we estimate corrected action value with
\begin{equation}
\textstyle
\hat{Q}^R_{t}=Q_t
        +\sum_{s \ge t}\gamma^{s-t^\prime}\left(\prod_{i=s}^{t}c_i\right)
        \delta_sQ,
\end{equation}
where $\delta_sQ=r_s+\gamma(\hat{V}_{s+1}-\alpha\KL_{s+1})-Q_s$ and  $Q_t=Q(a_{t},z_{t},x_{t})$.
$\hat{V}_s=\EE_{\pi(a|\epsilon_t,x_t)}[Q(a,z_{t},x_{t})]$ is estimated bootstrap value, $\KL_{s}=\KL\left[\pi^H(z|x_{s})\|\pi^H_{0}(z|x_{s})\right]+\KL\left[\pi^L(a|z_s,x_{s})\|\pp^L(a|z_s,x_{s})\right]$ and $\gamma$ is discount.
$c_{i}=\lambda \min\left(\frac{\pi(a_{i}|\epsilon_{i},x_{i})}{\mu(a_i|x_i)}\right)$ is truncated importance weight called \textit{traces}.

There are, however, a few notable details that we adapt for our method.
Firstly, we do not use the latent $z_t$ sampled from behaviour policies in actors. This is possible because the latent does not affect the environment directly. Instead, we consider the behavior policy as $\mu(a|x)$, which does not depend on latents. This approach is useful since we do not need to consider the importance weight with respect to the HL policy, which might introduce additional variance in the estimator.
Another detail is that the KL term at step $s$ is not considered in $\delta_sQ$ because the KL at step $s$ is not the result of action $a_s$.
Instead, we introduce close form KL at step $s$ as a loss to compensate for this.
The pseudocode for the resulting algorithm is illustrated in Algorithm~\ref{alg:r2h2}.

\setcounter{algorithm}{1}

\begin{algorithm*}[tb]
   \caption{SVG(0)~\cite{heess2015learning} with experience replay for hierarchical policy}
   \label{alg:r2h2}
\begin{algorithmic}
   \STATE Flat policy: $\pi_\theta(a_t|\epsilon_t,x_t)$ with parameter $\theta$
   \STATE HL policy: $\pi^H_\theta(z_t|x_t)$, where latent is sampled by reparameterization $z_t=f^H_\theta(x_t, \epsilon_t)$
   \STATE Default policies: $\pi^H_{0,\phi}(z_t|x_t)$ and $\pi^L_{0,\phi}(a_t|z_t,x_t)$ with parameter $\phi$
   \STATE Q-function: $Q_\psi(a_t,z_t,x_t)$ with parameter $\psi$
   \STATE Initialize target parameters $\theta^\prime \leftarrow \theta$,
   \hspace{0.2cm} $\phi^\prime \leftarrow \phi$, \hspace{0.2cm} $\psi^\prime \leftarrow \psi$.
   \STATE Target update counter: $c \leftarrow 0$
   \STATE Target update period: $P$
   \STATE Replay buffer: $\mathcal{B}$
   \REPEAT
   \FOR{$t=0, K, 2K, ... T$}
   \STATE Sample partial trajectory $\tau_{t:t+K}$ with action log likelihood $l_{t:t+K}$ from replay buffer $\mathcal{B}$:\\ \hspace{0.2cm}$\tau_{t:t+K}=(s_t,a_t,r_t,...,r_{t+K})$,\\
   \hspace{0.2cm}$l_{t:t+K}=(l_t,...,l_{t+K})=(\log\mu(a_t|x_t),...,\log\mu(a_{t+K}|x_{t+K}))$
   \STATE Sample latent: $\epsilon_{t^\prime} \sim \rho(\epsilon)$, $z_{t^\prime}=f^H_\theta(x_{t^\prime},\epsilon_{t^\prime})$
   \STATE Compute KL: $\hat{\KL}_{t^\prime}=\KL\left[\pi^H_\theta(z|x_{t^\prime})\|\pi^H_{0,{\phi^\prime}}(z|x_{t^\prime})\right]+\KL\left[\pi^L_\theta(a|z_{t^\prime},x_{t^\prime})\|\pi^L_{0,\phi^\prime}(a|z_{t^\prime},x_{t^\prime})\right]$
   \STATE Compute KL for Distillation:\\
   \hspace{0.2cm}$\hat{\KL}^\mathcal{D}_{t^\prime}=\KL\left[\pi^H_\theta(z|x_{t^\prime})\|\pi^H_{0,{\phi}}(z|x_{t^\prime})\right]+\KL\left[\pi^L_\theta(a|z_{t^\prime},x_{t^\prime})\|\pi^L_{0,\phi}(a|z_{t^\prime},x_{t^\prime})\right]$
   \STATE Compute action entropy:
   $\hat{\text{H}}_{t^\prime}=\EE_{\pi_\theta(a|\epsilon_{t^\prime},x_{t^\prime})}[\log\pi_\theta(a|\epsilon_{t^\prime},x_{t^\prime})]$
   \STATE Estimate bootstrap value:
   $\hat{V}_{t^\prime}
        =\EE_{\pi_\theta(a|\epsilon_{t^\prime},x_{t^\prime})}
        \left[Q_{\psi^\prime}(a,z_{t+K},x_{t+K})\right]-\alpha\hat{\KL}_{t+K}$
   \STATE Estimate traces~\cite{munos2016safe}:
   $\hat{c}_{t^\prime}=\lambda \min\left(\frac{\pi_\theta(a_{t^\prime}|\epsilon_{t^\prime},x_{t^\prime})}{l_{t^\prime}}\right)$
   \STATE Apply Retrace to estimate Q targets~\cite{munos2016safe}:\\
   \hspace{0.2cm}
   $\hat{Q}^R_{t^\prime}=Q_{\psi^\prime}(a_{t^\prime},z_{t^\prime},x_{t^\prime})
        +\sum_{s \ge t^\prime}\gamma^{s-t^\prime}\left(\prod_{i=s}^{t^\prime}\hat{c}_i\right)
        \left(r_s+\gamma\left(\hat{V}_{s+1}-\alpha\hat{\KL}_{s+1}\right)-Q_{\psi^\prime}(a_s,z_s,x_s)\right)$
   \STATE Policy loss: $\hat{L}_\pi=\sum_{i=t}^{t+K-1}\EE_{\pi_\theta(a|\epsilon_{i},x_{i})}Q_{\psi^\prime}(a,z_{i},x_{i})-\alpha\hat{\KL}_i + \alpha_\text{H}\hat{\text{H}}_i$
   \STATE Q-value loss: $\hat{L}_Q=\sum_{i=t}^{t+K-1}\|\hat{Q}^R_i-Q_\psi(a,z_{i},x_{i})\|^2$
   \STATE Default policy loss: $\hat{L}_{\pi^H_0}=\sum_{i=t}^{t+K-1}\hat{KL}_i^\mathcal{D}$
   \STATE
   $\theta \leftarrow \theta+\beta_\pi\nabla_\theta\hat{L}_\pi$ \hspace{0.3cm}
   $\phi \leftarrow \phi+\beta_{\pi^H_0}\nabla_\phi\hat{L}_{\pi^H_0}$
   \hspace{0.3cm}
   $\psi \leftarrow \psi-\beta_{Q}\nabla_\psi\hat{L}_{Q}$
   \STATE Increment counter $c \leftarrow c+1$
   \IF{$c > P$}
   \STATE Update target parameters $\theta^\prime \leftarrow \theta$,
   \hspace{0.2cm} $\phi^\prime \leftarrow \phi$, \hspace{0.2cm} $\psi^\prime \leftarrow \psi$
   \STATE $c \leftarrow 0$
   \ENDIF
   \ENDFOR
   \UNTIL
\end{algorithmic}
\end{algorithm*}

\subsection{Discrete action space}

For discrete control problems, we use distributed learning with V-trace~\cite{espeholt2018impala} off-policy correction.
Similarly to the distributed learning setup in continuous control, behaviours policies in multiple
actors are used to collect trajectories and a single learner is
used to optimize model parameters.
The learning algorithm is almost identical to~\cite{espeholt2018impala}, but there are details that need to be considered mainly because of hierarchy with stochastic latent variable and temporal abstraction.
Using negative KL as reward introduces another complication as well.

We consider optimizing objective with infrequent latent
\begin{equation}
\begin{split}
\textstyle
\mathcal{L}(&\qq, \pp)
\geq \\
&\textstyle \EE_{\traj}\left[
\sum_{t\ge1} \gamma^t r(s_t,a_t)
    - \alpha\gamma^t\mathds{1}_p(t)\KL(z_t|x_t) \right],
\label{eq:final_objective}
\end{split}
\end{equation}
where $\mathds{1}_p(t)$ is the indicator function whose value is $1$ if $t \hspace{-0.2cm} \mod p \equiv 1$ with period $p$.
This lower bound will be discussed later in Appendix~\ref{appendix:sec:sec:discount_lower_bound_derivation}.
This infrequent latent case is used for discrete action space experiment, by defining period $p$ to be equal to the effective step size of the body.

We learn latent conditional value function $V_\psi(z_t, x_t)$ and reparameterized flat policy $\pi_\theta(a_t|\epsilon_t, x_t)$.
V-trace target is computed as follows
\begin{equation}
\textstyle
v_s = V_\psi^s +
\sum_{t\geq s}\gamma^{t-s}\left(\prod_{i=s}^{t-1}c_i\right)
    \delta_tV,
\end{equation}
where $\delta_tV=\rho_t(r_t + \gamma (V_\psi^{t+1}-\KL^p_{t+1}) - V_\psi^{t})$, $\KL^p_t=\mathds{1}_p(t)\KL\big[\pi^H_\theta(z|x_t) \| \pi^h_{0,\phi}(z|x_t)\big]$, and $V_\psi^t \defeq V_\psi(z_t,x_t)$ is bootstraped value at time step $t$.
Importance weights are computed by $c_i\defeq \min(\bar{c},w_i)$ and $\rho_i\defeq \min(\bar{\rho},w_i)$, where $w_i=\frac{\pi_\theta(a_i|\epsilon_i,x_i)}{\mu(a_i|x_i)}$.
$\bar{c}$ and $\bar{\rho}$ are truncation coefficient identical to ones from original V-trace paper~\cite{espeholt2018impala}.
Note that here we ignore latent sampled by behaviour policy and just consider states and actions from the trajectory. 
As discussed in Appendix~\ref{sec:sec:alg_continuous}, we sample latent on-policy and this helps avoiding additional variance introduced with importance weight for HL policy.

Computed V-trace target is used for training both policy and value function with actor-critic algorithm.
For training policy, we use policy gradient defined as
\begin{equation}
\textstyle
\sum_{t \geq 1}
\rho_t\nabla_{\theta}
    \log\pi_\theta( a_t|z_t,x_t)\delta_tv
\end{equation}
where $\delta_tv=\hat{r}_t + \gamma (v_{t+1}-\KL^p_{t+1}) - V_\psi^t$ and $V_\psi^t = V_\psi(z_t,x_t)$.
We optimize negative KL for time step $t$ by adding an analytic loss function for HL policy $\pi^H_\theta(z|x_t)$ and default policy $\pi^H_\phi(z|x_t)$
\begin{equation}
\textstyle
\sum_{t\geq1}\nabla_{\theta,\phi} \mathds{1}_p(t)\KL[\pi^H_\theta(z|x_t) \|\pi^H_{0,\phi}(z|x_t)]
\end{equation}
For training value function, perform gradient descent over $l2$ loss
\begin{equation}
\textstyle
\sum_{t \geq 1}
(v_t-V^t_\psi)
\nabla_{\psi} V_\psi(z_t,x_t).
\end{equation}
Additionally we include an action entropy bonus to encourage exploration~\cite{mnih2016asynchronous}
\begin{equation}
\textstyle
\sum_{t \geq 1} \nabla_\theta \text{H}[\pi_\theta(a|\epsilon_t,x_t)],
\end{equation}
where $\text{H}[\cdot]$ is close form entropy.
We optimize the gradients from all four objectives jointly.
Unlike continuous control, we do not maintain target parameters separately for the discrete action space experiments.
\section{Derivations}
\label{appendix:derivations}
This section includes derivations not described in the main paper.

\subsection{Upper bound of KL divergence}
\label{appendix:sub:kl_upper_bound}

The upper bound of KL divergence in eq.~\eqref{eq:kl_upper_bound} of the main paper is derived as
\begin{align}
\KL(a_t|x_t)
\le& \KL(a_t|x_t) + \EE_{\qq(a_t|x_t)}[\KL(z_t|a_t,x_t)] \nonumber \\
=&
\EE_{\qq(a_t|x_t)}[
\textstyle \log \frac{\qq(a_t|x_t)}{\pp(a_t|x_t)}] \nonumber\\
 &\textstyle +\EE_{\qq(a_t|x_t)}[
\EE_{\qq(z_t|a_t,x_t)}[\textstyle \log \frac{\qq(z_t|a_t,x_t)}{\pp(z_t|a_t,x_t)}]
\nonumber \\
=&
\EE_{\qq(a_t,z_t|x_t)}[
\textstyle \log \frac{\qq(a_t,z_t|x_t)}{\pp(a_t,z_t|x_t)}]
=
\KL(z_t,a_t|x_t) \nonumber \\
=&
\EE_{\qq(z_t|x_t)}[
\textstyle \log \frac{\qq(z_t|x_t)}{\pp(z_t|x_t)}] \nonumber \\
 &\textstyle +\EE_{\qq(z_t|x_t)}[
\EE_{\qq(a_t|z_t,x_t)}[\textstyle \log \frac{\qq(a_t|z_t,x_t)}{\pp(a_t|z_t,x_t)}]
\nonumber \\
=&\KL(z_t|x_t) + \EE_{\qq(z_t|x_t)}[\KL(a_t|z_t,x_t)]
\end{align}
with the last expression being tractably approximated using Monte Carlo sampling. Note that:
\begin{align*}
\KL(z_t|x_t) &= \KL\big(\pi^H(z_t|x_t)\|\pi^H_0(z_t|x_t)\big) \\
\KL(a_t|z_t,x_t) &= \KL\big(\pi^L(a_t|z_t,x_t)\|\pi^L_0(a_t|z_t, x_t)\big)
\end{align*}

\subsection{Trajectory based derivation of lower bound}
\label{appendix:sec:sec:discount_lower_bound_derivation}

This section derives lower bound of eq.~\eqref{eq:objective:KL_regularized}~in the main paper.
We consider high level policy and default policy, whose latent are sampled every $p$ steps.
As in the lower bound in eq.~\eqref{eq:full_objective} of the main paper, we consider a case where a latent is shared between the policy and the prior.
We derive lower bound in trajectory level while considering both the period of high level action $p$ and the discount $\gamma$.
Here we introduce notation for trajectory including latent $\eta=(s_1,z_1,a_1,...)$, not including latent $\tau=(s_1,a_1,...)$, and including only the sequence of latent $\zeta=(z_1,z_{1+p},...)$.
In this section, we derive the following lower bound
\begin{equation}
\begin{split}
\textstyle
\mathcal{L}(\qq, \pp) = \textstyle \EE_{\traj} \big[ 
&\textstyle\sum_{t\ge1}  \gamma^t r(s_t,a_t) 
      - \alpha \gamma^t \KL(a_t|x_t) \big]\\
 \textstyle \geq \textstyle \EE_{\eta}\big[
&\textstyle\sum_{t\ge1} \gamma^t r(s_t,a_t)
    - \alpha\gamma^t\mathds{1}_p(t)\KL(z_t|x_t) \\
&\textstyle - \alpha\gamma^t\KL(a_t|z_{p(t)},x_t)
    \big],
\label{eq:kl_trajectory_lower_bound}
\end{split}
\end{equation}
where $\EE_{\traj}[\cdot]$ is taken with respect to the  distribution over trajectories defined by the agent policy and system dynamics: $p(s_1) \prod_{t\ge1} \qq(a_t | x_t) p(s_{t+1} | s_t, a_t)$.
$\EE_{\eta}[\cdot]$ is taken with respect to the distribution over trajectories of hierarchical policy including latent: $p(s_1)\prod_{t\ge1}\pi(a_t|z_{p(t)},x_t)\pi(z_t|x_t)^{\mathds{1}_p(t)}p(s_{t+1}|s_t,a_t)$. $z_{p(t)}$ is the latest latent sample before time step $t$ based on the period $p$ and $\mathds{1}_p(t)$ is the indicator function whose value is $1$ if $t \hspace{-0.2cm} \mod p \equiv 1$ with period $p$.

Note that these equations are composed of reward maximization and KL regularization term.
We could show equality in reward maximization, and derive lower bound with respect to the KL regularization term.
Therefore, we will present equality and inequality with respects to these two terms separately.

\subsubsection{Equality in Reward Maximization}

The equality of reward maximization term holds as follows
\begin{equation}
\begin{split}
\textstyle
\bbE_{\pi_\tau}&\textstyle\big[\sum_{t\geq1}\gamma^t r(s_t, a_t)\big]\\
&\textstyle = \int_\tau\pi(\tau)\big[\sum_{t\geq1}\gamma^t r(s_t, a_t)\big] d\tau \\
&\textstyle = \int_\tau \int_{\xi}\pi(\eta)d\xi\big[\sum_{t\geq1}\gamma^t r(s_t, a_t)\big] d\tau\\
&\textstyle = \int_\tau \int_{\xi}\pi(\eta)\big[\sum_{t\geq1}\gamma^t r(s_t, a_t)\big] d\xi d\tau \\
&\textstyle = \int_\eta\pi(\eta)\big[\sum_{t\geq1}\gamma^t r(s_t, a_t)\big] d\eta\\
&\textstyle = \bbE_{\pi_\eta}\big[\sum_{t\geq1}\gamma^t r(s_t, a_t)\big].
\end{split}
\end{equation}

\subsubsection{Lower Bound of KL regularization on trajectory}

We can show inequality in eq.~\eqref{eq:kl_trajectory_lower_bound} by deriving it from KL regularization term.
We first derive lower bound of trajectory level KL regularization without considering discount.
\begin{equation}
\begin{split}
\textstyle
\int_\tau \pi(\tau) &\textstyle \log \frac{\pi_0(\tau)}{\pi(\tau)}d\tau\\
&\textstyle =\int_\tau \pi(\tau) \big[\log \int_\xi \frac{\pi_0(\tau, \xi)}{\pi(\tau)}d\xi \big]d\tau\\
&\textstyle =\int_\tau \pi(\tau) \big[\log \int_\xi \pi(\xi|\tau)\frac{\pi_0(\tau, \xi)}{\pi(\tau)\pi(\xi|\tau)}d\xi \big]d\tau\\
&\textstyle \geq \int_\tau \pi(\tau) \big[\int_\xi \pi(\xi|\tau) \log \frac{\pi_0(\tau, \xi)}{\pi(\tau)\pi(\xi|\tau)}d\xi \big]d\tau\\
&\textstyle =\int_\eta \pi(\eta) \log \frac{\pi_0(\eta)}{\pi(\eta)}d\eta,
\label{eq:trajectory_lower_bound}
\end{split}
\end{equation}
where the inequality holds by Jensen's inequality.

\subsubsection{Lower Bound of KL Regularization with Discount}

To derive lower bound with discount, we first introduce the following equation to handle discount in the derivation.
\begin{equation}
\begin{split}
\textstyle
\sum_{t\geq1}^{T}\gamma^t a_t = & \textstyle \sum_{t\geq1}^{T-1}\left[(1-\gamma)\gamma^t\sum_{u\geq1}^{t}a_u\right]\\
&\textstyle+ \gamma^T\sum_{t\geq1}^T a_t.
\label{eq:unroll_discount}
\end{split}
\end{equation}
This equality is useful to derive lower bound from summation without discounting ( $\sum_{u\geq1}^ta_u$) and then recombine it with the discounting terms.

To derive lower bound with respect to KL regularization with discount, we first rewrite KL regularization term as a trajectory based KL term.
As $\alpha$ is a constant we will ignore it in the derivation, but it is straightforward to include it.
We could turn KL regularization term as a trajectory based KL
\begin{equation}
\begin{split}
\textstyle
\bbE_{\pi_\tau} &\textstyle \big[\sum_{t\geq1}^T-\KL\big(a_t|x_t\big)
\big]\\
&\textstyle =-\int_\tau \pi(\tau) \big[\sum_{t\geq1}^T\KL\big(a_t|x_t\big)\big]d\tau \\
&\textstyle = \int_\tau \pi(\tau) \big[\sum_{t\geq1}^T\int_{a_t}\pi(a_t|x_t)\log \frac{\pi_0(a_t|x_t)}{\pi(a_t|x_t)}da_t \big]d\tau\\
&\textstyle = \int_\tau \pi(\tau) \big[\sum_{t\geq1}^T\log \frac{\pi_0(a_t|x_t)}{\pi(a_t|x_t)} \big]d\tau\\
&\textstyle =\int_\tau \pi(\tau) \log \prod_{t\geq1}^T\frac{\pi_0(a_t|x_t)}{\pi(a_t|x_t)}d\tau\\
&\textstyle =\int_\tau \pi(\tau) \log \frac{\pi_0(\tau)}{\pi(\tau)}d\tau. \\
\label{eq:kl_regularization_to_trajectory}
\end{split}
\end{equation}

We use eqs.~\eqref{eq:unroll_discount} and \eqref{eq:kl_regularization_to_trajectory} to rewrite discounted KL regularization as a trajectory based equation.
\begin{equation}
\begin{split}
\textstyle
&\bbE_{\pi_\tau}
\big[
    \sum_{t\geq1}^T
    \textstyle - \gamma^t\KL\big(a_t|x_t\big)
\big] \\
&\textstyle\hspace{0.5cm} = \bbE_{\pi_\tau}
\big[\textstyle -\sum_{t\geq1}^{T-1}\big[(1-\gamma)\gamma^t\sum_{u\geq1}^{t}\KL\big(a_u|x_u\big)\big] \\
&\textstyle\hspace{1.0cm}- \gamma^T\sum_{t\geq1}^T \KL\big(a_t|x_t\big)
\big] \\
&\textstyle\hspace{0.5cm} = 
    -\sum_{t\geq1}^{T-1}\big[\textstyle (1-\gamma)\gamma^t
        \bbE_{\pi_\tau} \big[
            \sum_{u\geq1}^{t}\KL\big(a_u|x_u\big)
        \big]
    \big] \\
    &\textstyle\hspace{1.0cm}- \gamma^T \bbE_{\pi_\tau} \big[
        \sum_{t\geq1}^T 
        \KL\big(a_t|x_t\big)
    \big] \\
&\textstyle\hspace{0.5cm} = 
    \sum_{t\geq1}^{T-1} \big[\textstyle (1-\gamma)\gamma^t
        \int_{\tau_t} \pi(\tau_t) \log \frac{\pi_0(\tau_t)}{\pi(\tau_t)}d\tau_t
    \big] \\
    &\textstyle\hspace{1.0cm}+ \gamma^T 
        \int_{\tau_T} \pi(\tau_T) \log \frac{\pi_0(\tau_T)}{\pi(\tau_T)}d\tau_T,
\end{split}
\end{equation}
where $\tau_t$ is trajectory until time step $t$.
We derive lower bound of this trajectory based representation using eq.~\eqref{eq:trajectory_lower_bound}.
\begin{equation}
\begin{split}
\textstyle
&\textstyle\sum_{t\geq1}^{T-1}\big[ (1-\gamma)\gamma^t
    \int_{\tau_t} \pi(\tau_t) \log \frac{\pi_0(\tau_t)}{\pi(\tau_t)}d\tau_t
\big] \\
&\textstyle\hspace{1.0cm}+ \gamma^T 
    \int_{\tau_T} \pi(\tau_T) \log \frac{\pi_0(\tau_T)}{\pi(\tau_T)}d\tau_T \\
&\textstyle \geq
\sum_{t\geq1}^{T-1}\big[ (1-\gamma)\gamma^t
    \int_{\eta_t} \pi(\eta_t) \log \frac{\pi_0(\eta_t)}{\pi(\eta_t)}d\eta_t
\big] \\
&\textstyle\hspace{1.0cm}+ \gamma^T 
    \int_{\eta_T} \pi(\eta_T) \log \frac{\pi_0(\eta_T)}{\pi(\eta_T)}d\eta_T,
\label{eq:appendix:z_lower_bound}
\end{split}
\end{equation}
To rearrange this trajectory based representation with KL regularization formulation, we use following equality
\begin{equation}
\begin{split}
\textstyle
\int_\eta &\textstyle \pi(\eta)  \log \frac{\pi_0(\eta)}{\pi(\eta)}d\eta\\
&\textstyle = \int_\eta \pi(\eta) 
    \log \prod_{t\geq1}^T
        \frac{
            \pi_0(a_t|z_{p(t)}, x_t)\pi_0(z_t|x_t)^{\mathds{1}_p(t)}}{
            \pi(a_t|z_{p(t)}, x_t)\pi(z_t|x_t)^{\mathds{1}_p(t)}}
    d\eta \\
&\textstyle= \int_\eta \pi(\eta) 
    \big[ \sum_{t\geq1}^T \log 
        \frac{
            \pi_0(a_t|z_{p(t)}, x_t)}{
            \pi(a_t|z_{p(t)}, x_t)}\\
            &\textstyle\hspace{1.0cm}+ \mathds{1}_p(t) \log
        \frac{
            \pi_0(z_t|x_t)}{
            \pi(z_t|x_t)}
    \big]
    d\eta \\
&\textstyle=
\int_\eta \pi(\eta)\big[
    \sum_{t \geq 1}^T
    - \mathds{1}_p(t)
        \KL\big(z_t|x_t\big)\\
        &\textstyle\hspace{1.0cm}-
        \KL\big(a_t|z_{p(t)},x_t\big)
\big] d\eta\\
&\textstyle=
\bbE_{\pi_\eta}\big[
    \sum_{t \geq 1}^T
    - \mathds{1}_p(t)
        \KL\big(z_t|x_t\big) -
        \KL\big(a_t|z_{p(t)},x_t\big)
\big],
\label{eq:lower_bound_rearrange}
\end{split}
\end{equation}
where $z_{p(t)}$ is the latest latent sample before time step $t$ based on the period $p$.

We rearrange eq.~\eqref{eq:appendix:z_lower_bound} based on eqs.~\eqref{eq:unroll_discount} and \eqref{eq:lower_bound_rearrange}.
\begin{equation}
\begin{split}
\textstyle
&\textstyle\sum_{t\geq1}^{T-1}\big[(1-\gamma)\gamma^t
    \int_{\eta_t} \pi(\eta_t) \log \frac{\pi_0(\eta_t)}{\pi(\eta_t)}d\eta_t
\big]\\
&\textstyle\hspace{1.0cm}+ \gamma^T 
        \int_{\eta_T} \pi(\eta_T) \log \frac{\pi_0(\eta_T)}{\pi(\eta_T)}d\eta_T \\
&\textstyle= 
-\sum_{t\geq1}^{T-1}\big[(1-\gamma)\gamma^t
    \bbE_{\pi_\eta}\big[\sum_{u \geq 1}^t
        \mathds{1}_p(u)\KL\big(z_u,x_u\big)\\
        &\textstyle\hspace{3.0cm}+ \KL\big(a_u|z_{p(u)},x_u\big)\big]
    \big]\\
    &\textstyle \hspace{0.5cm} - \gamma^T 
        \bbE_{\pi_\eta}\big[\sum_{t \geq 1}^T
            \mathds{1}_p(t)\KL\big(z_t|x_t\big)+\KL\big(a_t|z_{p(t)},x_t\big)\big] \\
&\textstyle= -\bbE_{\pi_\eta}\big[
    \sum_{t\geq1}^{T-1}\big[(1-\gamma)\gamma^t
        \sum_{u \geq 1}^t
            \mathds{1}_p(u)\KL\big(z_u|x_u\big)\\
            &\textstyle\hspace{3.0cm}+ \KL\big(a_u|z_{p(u)},x_u\big)
        \big]
    \\
    & \textstyle \hspace{0.5cm} + \gamma^T 
            \sum_{t \geq 1}^T
                \mathds{1}_p(t)\KL\big(z_t|x_t\big) +\KL\big(a_t|z_{p(t)},x_t\big)
    \big] \\
&\textstyle= \bbE_{\pi_\eta}\big[
    \sum_{t\geq1}^{T}
        -\mathds{1}_p(t)\gamma^t\KL\big(z_t|x_t\big) 
        -\gamma^t \KL\big(a_t|z_{p(t)},x_t\big)
\big].
\end{split}
\end{equation}
By combining all results, we obtain the following inequality.
\begin{equation}
\begin{split}
\textstyle
\mathcal{L}(\qq, \pp) = \textstyle \EE_{\traj} \big[ 
&\textstyle\sum_{t\ge1}  \gamma^t r(s_t,a_t) 
      - \alpha \gamma^t \KL(a_t|x_t) \big]\\
 \textstyle \geq \textstyle \EE_{\eta}\big[
&\textstyle\sum_{t\ge1} \gamma^t r(s_t,a_t)
    - \alpha\gamma^t\mathds{1}_p(t)\KL(z_t|x_t)\\
    &\textstyle- \alpha\gamma^t\KL(a_t|z_{p(t)},x_t)
    \big].
\label{eq:trajectory_lower_bound_again}
\end{split}
\end{equation}

\begin{figure*}[t]
    \centering
    \subfigure[Learning curves with quasi on-policy training.]{\includegraphics[width=.4\linewidth]{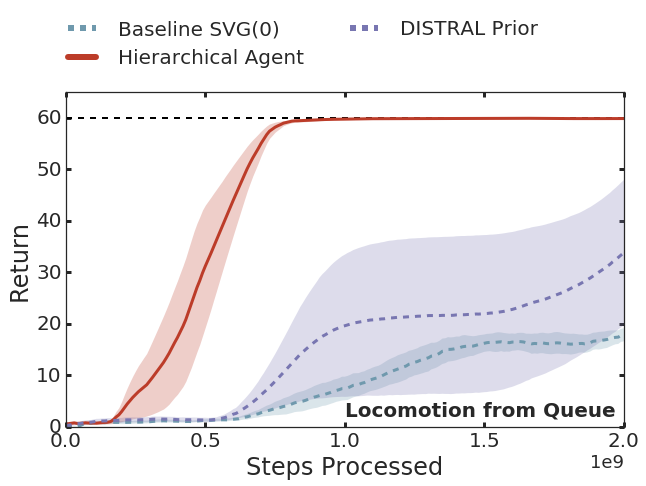}} 
    \label{fig:result_from_queue}
    \subfigure[Learning curves with a single actor.]{\includegraphics[width=.4\linewidth]{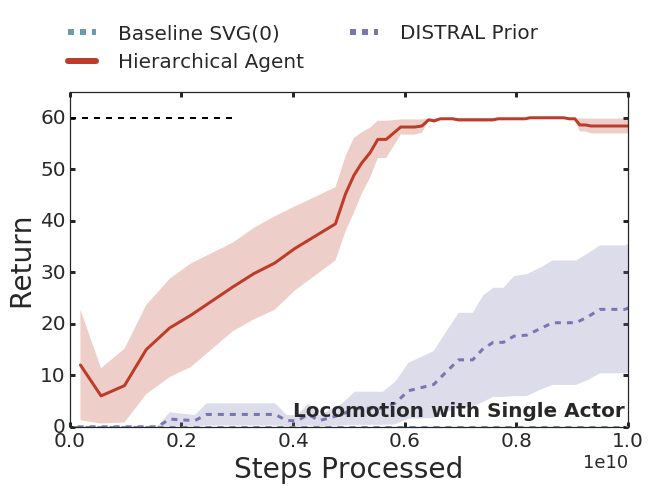}} 
    \label{fig:result_from_single_actor}
    \vspace{-0.2cm}
    \caption{\textbf{Experiments with alternative training regimes.} (a) Locomotion with Ant. AR-1 process. (b) Locomotion with Ant. AR-1 process.}
    \vspace{-0.2cm}
    \label{fig:alternative_training_regimes}
\end{figure*}

\begin{figure}[t!]
    \raggedright
    \vspace{-0.2cm}
    \includegraphics[width=1.0\linewidth]{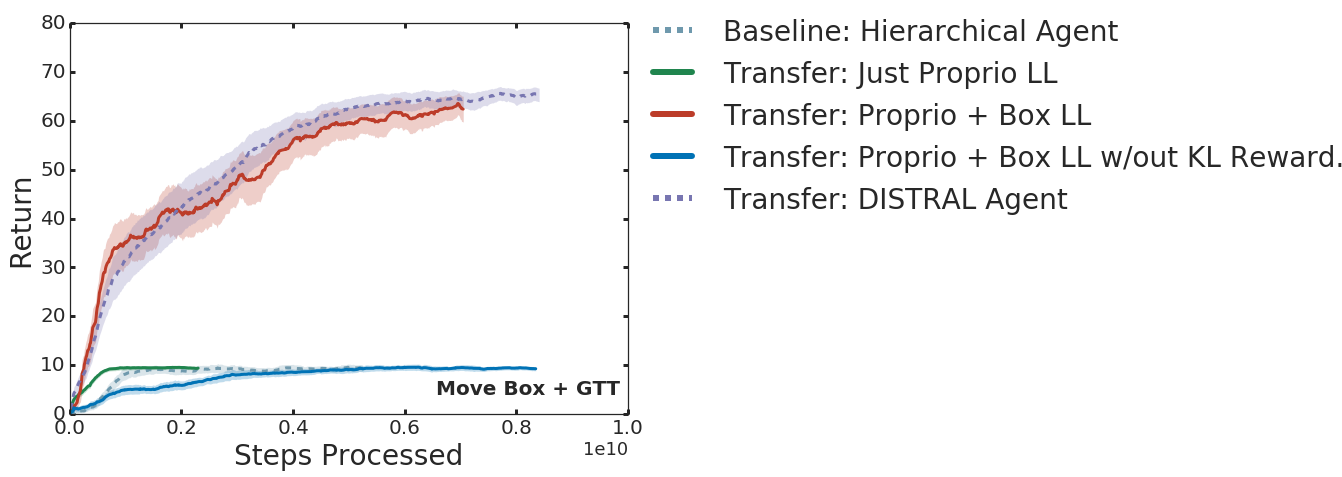}
    \vspace{-0.4cm}
    \caption{
        \textbf{Analysis on task transfer results.} Transfer from Combined task (OR) to Combined task (AND) with Ant and AR-1 process.
    }
    \vspace{-0.3cm}
    \label{fig:task_transfer_continuous_LL_input}
\end{figure}

\section{Additional Experimental Results}

\subsection{Alternative training regimes}

In the main paper, we present results based on learning speed with respect to the number of time steps processed by learner in distributed learning setup.
Note that the number of time steps processed by the learner does not necessarily correspond to the number of collected trajectory time steps because of the use of experience replay, which allows to learning to proceed regardless of the amount of collected trajectories. We also experimented with two alternative training regimes to ensure that the speedup results reported are consistent.
In Figure~\ref{fig:alternative_training_regimes}a, we compare the learning curves for our method against the SVG-0 and DISTRAL baselines in a quasi on-policy training regime similar to that of \cite{espeholt2018impala}. In Figure~\ref{fig:alternative_training_regimes}b, we perform a similar comparison in the original replay based off-policy training regime but with a single actor generating the data. In both cases, our method learns faster than both baselines.

\subsection{Information asymmetry in task transfer}

Figure~\ref{fig:task_transfer_continuous_LL_input} illustrates the necessity of information asymmetry and KL regularization during transfer. Here we train the agent on the Move box Or Go to Target task with different information given to the LL and then transfer to Move Box and GTT. Therefore the distributions of the trajectories during training and those required for transfer should be similar. 
As the figure shows, we observe successful transfer only when the box position is given to the LL controller and KL regularization is used during transfer.
Failed approaches usually converge to suboptimal policies, where the agent succeeds on the go to target task, but cannot move the box appropriately.
In this transfer scenario, giving box position as input to LL controller is one way to specify inductive bias, which turns out to be useful to move box appropriately in target task. Interestingly, the unstructured DISTRAL prior performs comparatively to our method in this experiment. We hypothesize that for tasks that take longer to learn, the benefit of the immediate parameter transfer in our approach is not as strong since this also leads to a fixed lower level behavior that cannot be adapted to the task. In this sense, the DISTRAL baseline is expected to be asymptotically stronger.

\begin{figure*}[t!]
    \centering
    \includegraphics[width=1.0\linewidth]{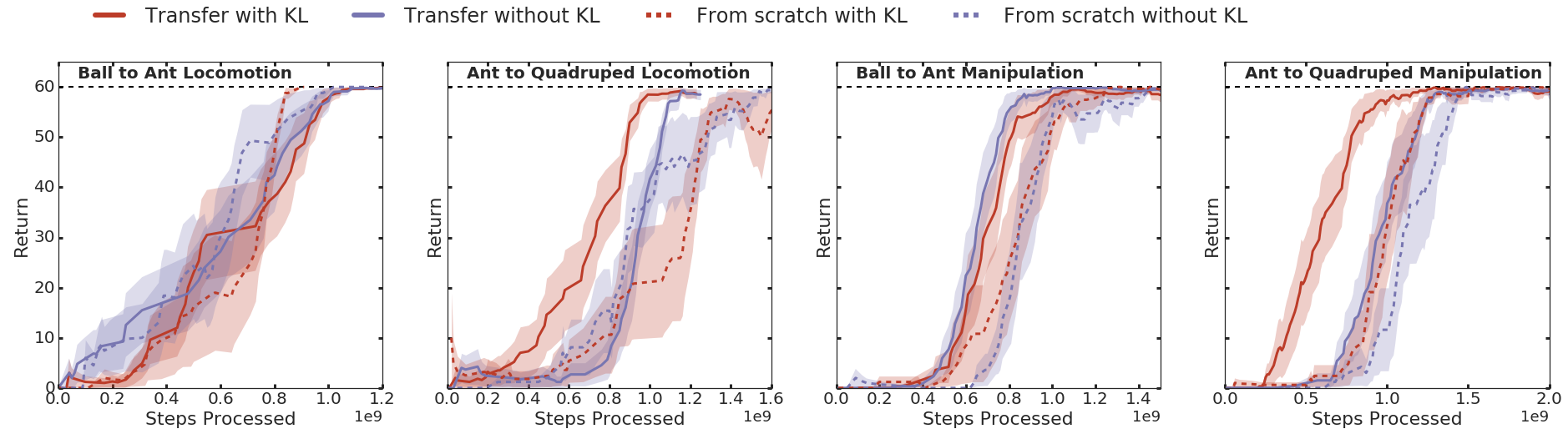}
    \vspace{-0.6cm}
    \caption{
        \textbf{Body transfer with the AR-1 Prior.} \textbf{Column 1}: Ball to Ant, Locomotion. \textbf{Column 2}: Ball to Ant, Manipulation (easy). \textbf{Column 3}: Ant to Quadruped, Locomotion. \textbf{Column 4}: Ant to Quadruped, Manipulation(easy).
    }
    \label{fig:body_transfer_continuous}
\end{figure*}

\subsection{Additional body transfer experiments}
\label{appendix:sub:additional_body_transfer}

We explore body transfer setup both in discrete and continuous environments.
We compare performance to learning the hierarchical policy from scratch and analyze the effects of the KL regularization.
The experimental setup in the continuous case is the same as before, and
Figure~\ref{fig:body_transfer_continuous} provides results for different types
of bodies and tasks. Generally transferring the HL component and relying on both the task reward and the KL term as a dense shaping reward signal for LL controller works best in these settings.

\begin{figure}
  \begin{center}
    \hspace{-0.2cm}\includegraphics[width=0.95\linewidth]{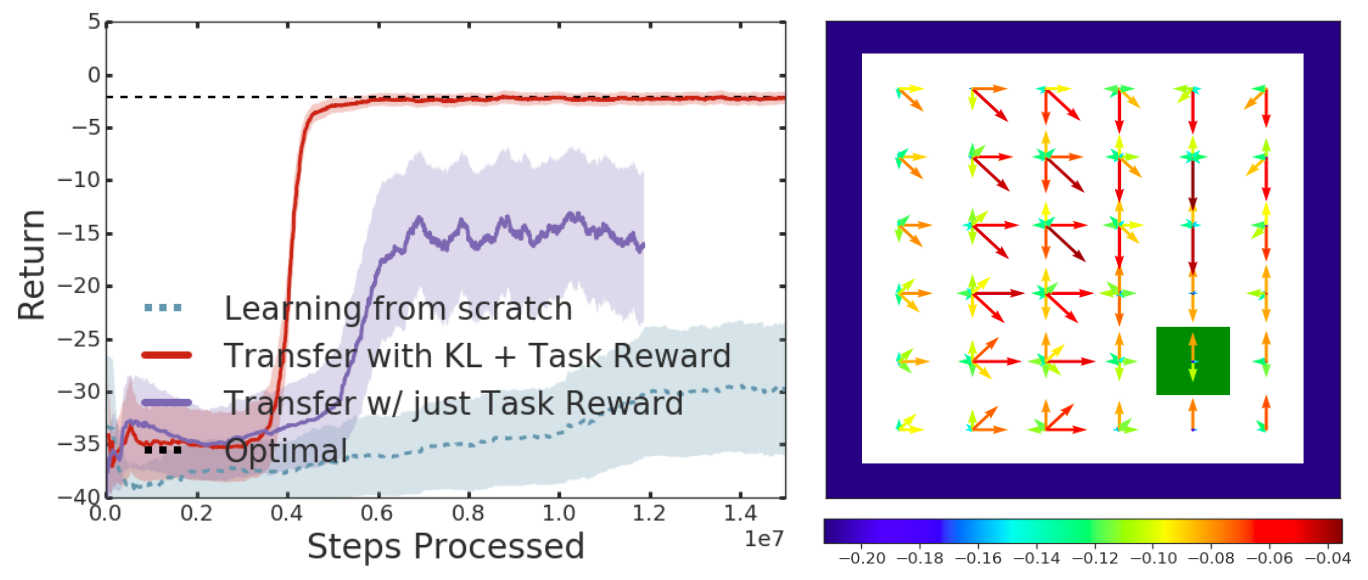}
  \end{center}
  \vspace{-0.2cm}
  \caption{
    \textbf{Body transfer in 2D grid world, AR-1 prior.} \textbf{Left} Transfer from 1-step to 8-step body. \textbf{Right} KL reward visualization. The size and the color of arrows denotes KL reward (negative KL) for corresponding agents' movement.
  }
  \vspace{-0.2cm}
  \label{fig:body_transfer_discrete}
\end{figure}


In the discrete case, we construct a \textbf{discrete go to target} task in a 2D grid world.
An agent and goal are randomly placed in an $8\times 8$ grid and the agent is rewarded for reaching the goal.
The body agnostic task observation is the global x, y coordinates of the agent and goal.
The different bodies in this case must take different numbers of actions to achieve an actual step in the grid. For instance the 4-step body needs to take 4 consecutive actions in the same direction to move forward by one step in that direction.
We assume that the latent $z_t$ is sampled every $n$ steps, where $n$ is the number of actions required to take a step. Details for models in which latent variables are sampled with a period $> 1$ are provided in Appendix~\ref{appendix:algorithm}. The environment is described in Appendix~\ref{appendix:environments}.

Figure~\ref{fig:body_transfer_discrete} illustrates the result for transfering behavior from the 1-step to the 8-step body with AR-learned prior. (We were only able to solve the challenging 8-step version through body transfer with a KL reward.)
In Figure~\ref{fig:body_transfer_discrete}, we visualize the negative KL divergence (KL reward) along the agent's movement in every location of the grid. The size and the colour of arrows denotes the expected KL reward. This illustrates that the KL reward forms a vector field that guides the agent toward the goal, which provides a dense reward signal when transferring to a new body.
This observation explains the gain from KL regularization, which can lead to faster learning and improve asymptotic performance.
\section{Environments}
\label{appendix:environments}
\subsection{Discrete control}

We construct a \textbf{discrete go to target} task in 2D grid world.
An agent and goal are randomly placed in an $8\times 8$ grid and the agent is given a reward of 1.0 for reaching the goal.
The episode terminates when the agent reaches the goal or after 400 time steps. Additionally, the agent receives a penalty of -0.1 for every time step and a penalty of -0.2 if it collides with the walls.
The body agnostic tasks observation is global x, y coordinate of agent and goal.

We consider a body that moves in any of 4 directions in the grid (up, down, left, right). 
We define different bodies based on their effective step size.
The effective step size is the number of consecutive movements required to make a single displacement in the grid.
Specifically, a body has 2 dimensional internal coordinate $[-n+1,n-1]^2$ with effective step size $n$.
Agent's action primarily affect the internal coordinate and it brings displacement to the external coordinate only if a value exceed its minimum or maximum. In this case, agent move 1 step in external coordinate and internal coordinate for the corresponding dimension is reset to $0$.
We denote different bodies with their step size (e.g. 1-step).
We assume that the latent $z_t$ is sampled every $n$ steps, where $n$ is the effective step size.

\subsection{Continuous control}

In this section, we describe detailed configuration of the continuous control tasks and bodies.

\subsubsection{Tasks}

\paragraph{Locomotion (Go to 1 of 3 targets)}
On a fixed $8\times8$ area, an agent and 3 targets are randomly placed at the beginning of episodes.
In each episode, one of the 3 targets are randomly selected, and the agent should reach the selected target within 400 time steps.
When the agent reach the selected target, the agent receives a reward of 60 and the episode terminates.
Egocentric coordinates of 3 targets and an onehot vector representing one of the 3 targets are provided as task observations.

\vspace{-0.4cm}

\paragraph{Locomotion (Gap)}
The task consists of a corridor with one gap in the middle. The length of the gap is chosen randomly at the start of each episode uniformly 
between 0.3 and 2.5. In order to successfully solve the task, the ant needs to jump across the gap and reach the end of the corridor. The ant 
recieves a reward proportional to its velocity at each timestep. The observations given to the agent include proprioceptive information regarding the
joint positions and velocities of the walker as well as the position and length of the gap for the current episode.

\vspace{-0.4cm}
\paragraph{Maniputation (Move 1 of N boxes to 1 of K targets)}
On a fixed $3\times3$ area, an agent, N boxes and K targets are randomly placed at the beginning of episodes.
In each episode, one of N boxes and one of the K targets are randomly selected, and the agent should move the box to the selected target within 400 time steps.
When the box is placed on the selected target, the agent receives a reward of 60 and the episode terminates.
Egocentric coordinates of K targets and 6 corner of the N boxes are provided as task observations with an onehot vector representing one of the K targets.

\vspace{-0.3cm}
\paragraph{Manipulation (gather)} 
On a fixed $3\times3$ area, an agent and 2 boxes are randomly locate at the beginning of episodes.
In each episode, the agent should gather all the boxes within 400 time steps so that the boxes are contacted to each other.
The agent receives a reward of 60 when it successfully gathers boxes.
Egocentric coordinates of 6 corners of boxes are provided as task observations.

\vspace{-0.3cm}
\paragraph{Combined task (Or)} 
This task is a combination of the go to 1 of 2 targets task and the move box to 1 of 2 targets task. On each episode one of the two tasks is randomly sampled so that with probability 0.5 the agent must go to one of the 2 targets and with probability 0.5 it must push a box to a specific target. The agent receives a reward of 60 for successfully completing the corresponding task. 
Egocentric coordinates of both targets as well as the corners of the box are provided as task observations along with an encoding describing the specific task instance.

\vspace{-0.3cm}
\paragraph{Combined task (And)}
This task contains the move box to 1 of 2 targets and go to 1 of 2 targets as two sub-tasks. In each episode, the walker must push the box to a target and then go to another target. A reward of 10 is awarded for each sub task that is completed and a bonus reward of 50 is awarded for completing both tasks.
Egocentric coordinates of both targets as well as the corners of the box are provided as task observations.

\begin{figure}[t!]
    \centering
    \includegraphics[width=1.0\linewidth]{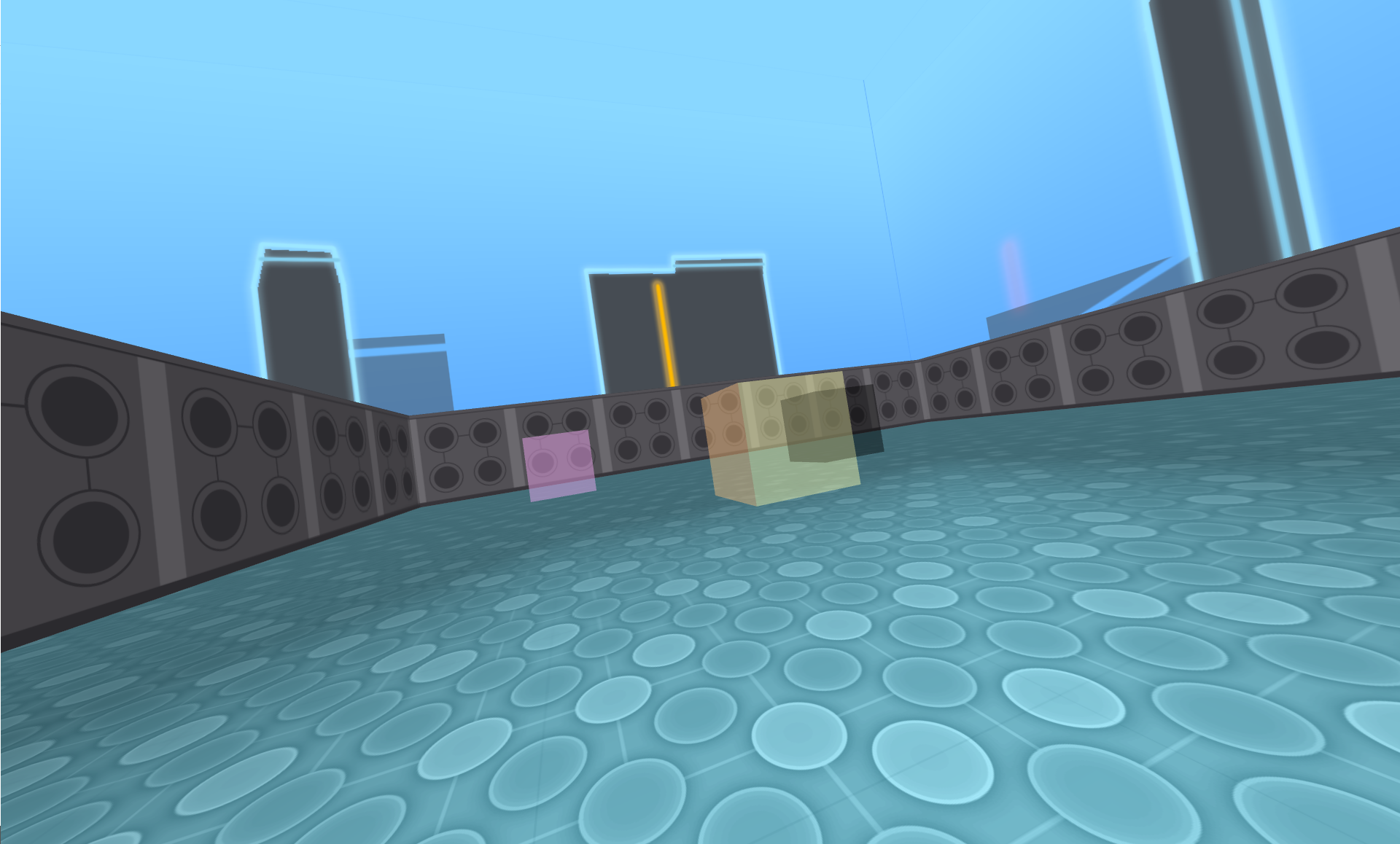}
    \vspace{-0.6cm}
    \caption{
        \textbf{Vision input for Go to target from vision.}
    }
    \label{fig:vision_gtt_vision}
    \vspace{-0.1cm}
\end{figure}

\vspace{-0.3cm}
\paragraph{Manipulation (vision)}
This task is identical to the Manipulation, but the task observation is egocentric vision.
Instead of egocentric target coordinates, the agent observes a $64\times64$ image captured by the agent's egocentric camera. 
The agent needs to recognize visual patterns of the targets and figure out the correct target (always colored black in this case).
Figure~\ref{fig:vision_gtt_vision} illustrates the egocentric visual observation before being rescaled to $64\times64$.

\subsubsection{Bodies}

We use three different bodies: Ball, Ant, and Quadruped.
Ball and Ant have been used in several previous works~\cite{heess2017emergence,xie2018transferring,galashov2018information}, and we introduce Quadruped as a more complex variant of the Ant.
The \textbf{Ball} is a body with 2 actuators for moving forward or backward, turning left, and turning right.
The \textbf{Ant} is a body with 4 legs and 8 actuators, which moves its legs to walk and to interact with objects.
The \textbf{Quadruped} is similar to Ant, but with 12 actuators.
Each body has different proprioception (proprio) as a body specific observation.

\begin{table}[]
\centering
\begin{tabular}{|p{2.3cm}|c|p{3.6cm}|}
\hline
 Task & Walker & LL information  \\
\hline
 Go to 1 of K Targets & Ant & Proprioception   \\
 \hline
 Move box to target & Ant &  Proprioception + Box  \\
 \hline
 2 Boxes and 2 Targets & Ball & Proprioception + Boxes + Targets \\
 \hline
 Move Box or Go to Target (I) & Ant & Proprioception \\
 \hline
 Move Box or Go to Target (II) & Ant & Proprioception + Box \\
 \hline
 Gather Boxes & Ball & Proprioception + Boxes + Targets \\
 \hline
\end{tabular}
\caption{Information provided to the lower level controller for each task.}
\label{tab:ll_info}
\end{table}

\subsubsection{Details of input to the Lower Level}
\ref{tab:ll_info} illustrates the information provided to the lower level for each of the tasks for the speedup and transfer settings considered in the main text. In these cases, the HL received full information.
For the tasks where the body was transferred, the LL was only given propioceptive information while the HL received all other information relevant to the task. 

\section{Experimental Settings}
\label{appendix:experimental_settings}

\subsection{General settings}
Throughout the experiments, we use 32 actors to collect trajectories and a single learner to optimize the model.
We plot average episode return with respect to the number of steps processed by the learner.
Note that the number of steps is different from the number of agent's interaction with environment, because the collected trajectories are processed multiple times by a centralized learner to update model parameters.
When learning from scratch we report results as the mean and standard deviation of average returns for 5 random seeds.
For the transfer learning experiments, we use 5 seed for the initial training, and then transfer all pretrained models to a new task, and train two new HL or LL controllers (two random seeds) per model on the transfer task. Thus,
in total, 10 different runs are used to estimate mean and standard deviations of the average returns.
Hyperparameters, including KL cost and action entropy regularization cost, are optimized on a per-task basis. More details are provided in Appendix~\ref{appendix:hyper_parameters}.

\subsection{DISTRAL with parameter sharing}
\label{appendix:sub:distral_share}

We introduced two baselines as variants of DISTRAL prior sharing parameters between the agent policy and the default policy.
\textbf{DISTRAL prior 2 cols} uses a 2-column architecture (see~\cite{teh2017distral}), where the default policy network is reused in combination with another network (column) to output the final policy distribution. In this configuration, the default policy network does not access to task information, but another network (column) access to the full information.
In \textbf{DISTRAL shared prior}, the policy network is reused to output the default policy distribution based on an additional branch on top of it. As default policy is constructed on policy network, information asymmetry is not used in this baseline.

\section{Hyper parameters}
\label{appendix:hyper_parameters}

Fully connected neural networks are used as function approximators for both the actor and the critic in the agent. In case of tasks with boxes, a separate common single layer MLP was used as a \textit{box encoder torso}. Separate processing networks were implemented for \textit{proprioception} for the baselines. For all tasks, multiple values were swept for actor networks and torso sizes with 5 random seeds each. ELU activations were used throughout. We use separate optimizers and learning rates for the actor and critic networks. For the hierarchical networks, fully connected MLP networks were used for the higher level and lower level policy cores. The relative contribution of the KL regularization to the reward was controlled by a \textit{posterior entropy cost} which we denote $\alpha$.

Below we provide the default hyperparameters used across tasks followed by the best parameters from the baselines and the hierarchical networks for each task.
\subsection{Default parameters}
\textit{Actor learning rate,} $\beta_{pi}$ = 1e-4.

\textit{Critic learning rate,} $\beta_{pi}$ = 1e-4.

\textit{DISTRAL baseline default policy learning rate,} $\beta_{pi}$ = 5e-4.

\textit{Target network update period} = 100.

\textit{DISTRAL policy target network update period} = 100.

\textit{Baseline Actor network:} MLP with sizes (200, 100).

\textit{Baseline Critic network:} MLP with sizes (400, 300).

\textit{DISTRAL baseline default policy network:} MLP with sizes (200, 100).

\textit{HL policy network:} MLP with sizes (200, 10).

\textit{LL policy network:} MLP with sizes (200, 100).

\textit{Box encoder network:} MLP with sizes (50).

\textit{Batch size:} 512.

\textit{Unroll length:} 10.

\textit{Entropy bonus,} $\lambda$ = 1e-4.

\textit{Posterior Entropy cost for HL,} $\alpha$ =  1e-3.

\textit{Posterior Entropy cost for DISTRAL default policy,} $\alpha$ = 0.01

\textit{Distillation cost for DISTRAL default policy,} $\alpha$ = 0.01

\textit{Number of actors:} 32

\subsection{Per-task parameters}
\textbf{Ant: Move Box to Target}

\textit{Entropy bonus,} $\lambda$ = 1e-3.

\textit{Policy network for Gaussian Prior:} MLP with sizes (200, 100).

\textit{Action Entropy cost for Isotropic Gaussian Prior:} =  0.

\textit{Action Entropy cost for AR-1 Prior:} =  0.

\textit{AR Parameter:} =  0.9

\textit{Posterior Entropy cost for AR-1 Prior,} $\alpha$ =  1e-4.

\textbf{Ant: Go to 1 of 3 Targets}

\textit{Action Entropy cost for DISTRAL default policy} =  0.

\textit{AR Parameter:} =  0.95

\textit{Distillation cost for DISTRAL default policy,} $\alpha$ = 1.0

\textbf{Ball: 2 Boxes to 2 Targets}

\textit{Distillation cost for DISTRAL default policy,} $\alpha$ = 0.1

\textit{Box encoder network:} MLP with sizes (100, 20).

\textit{HL policy network for AR-Learned Prior:} MLP with sizes (200, 4).

\textbf{Ant: Move Box to 1 of 3 Targets}

\textit{AR Parameter for baseline:} =  0.9

\textit{Policy network for AR-1 Transfer Prior:} MLP with sizes (200, 10).

\textit{Policy network for Gaussian Transfer Prior:} MLP with sizes (200, 4).

\textit{Policy network for AR-Learned Transfer Prior:} MLP with sizes (200, 4).

\textit{Posterior Entropy cost for AR-Learned Transfer Prior,} $\alpha$ =  1e-2.

\textit{Action Entropy cost for AR-Learned Transfer Prior:} =  1e-4.

\textbf{Ball: Gather boxes}

\textit{Policy network for Gaussian Transfer Prior:} MLP with sizes (200, 4).

\textit{Policy network for AR-Learned Transfer Prior:} MLP with sizes (200, 4).

\textit{DISTRAL Actor learning rate,} $\beta_{pi}$ = 5e-4.

\textit{DISTRAL default policy distillation cost} = 0.01.

\textbf{Ant: Move Box and Go to Target}

\textit{Policy network for AR-1 Prior:} MLP with sizes (200, 10).

\textbf{Ball to Ant: Go to Target}

Unless otherwise specified, for all the body transfer tasks an \textit{action entropy cost} of 1e-4 worked best across tasks.

\textit{HL Policy network for agent from scratch}: MLP with sizes (100, 4)

\textbf{Ball to Ant: Move Box to Target}

\textit{HL Policy network for agent from scratch}: MLP with sizes (100, 4)

\textit{Posterior entropy cost for transfer agent}: 1e-5

\textbf{Ant to Quadruped: Move Box to Target}

\textit{Posterior entropy cost for transfer agent}: 1e-2

\textit{Action entropy cost for transfer agent}: 1e-5

\textbf{Ant to Quadruped: Go To Target}:

\textit{HL Policy network for agent from scratch}: MLP with sizes (20)

\textbf{Ant to Quadruped: Go To Target From Vision}:

\textit{HL Policy network}: Residual Network with an embedding size of (256,)

\end{document}